\documentclass[10pt,twocolumn,letterpaper]{article}

\usepackage{cvpr}              

\usepackage[final]{graphicx}
\usepackage{amsmath}
\usepackage{amssymb}
\usepackage{booktabs}
\usepackage[T1]{fontenc}
\usepackage{cite}
\usepackage{amsmath}
\usepackage{cuted}
\usepackage{lipsum}
\usepackage{float}
\usepackage{multirow}
\usepackage[accsupp]{axessibility} 
\usepackage[pagebackref,breaklinks,colorlinks]{hyperref}

\usepackage[capitalize]{cleveref}
\crefname{section}{Sec.}{Secs.}
\Crefname{section}{Section}{Sections}
\Crefname{table}{Table}{Tables}
\crefname{table}{Tab.}{Tabs.}

\def\cvprPaperID{9817} 
\def\confName{CVPR}
\def\confYear{2023}

\begin{document}

\title{Multiclass Confidence and Localization Calibration for Object Detection}





\author{Bimsara Pathiraja
~~~
Malitha Gunawardhana
~~~
Muhammad Haris Khan
~~~
\\
Mohamed bin Zayed University of Artificial Intelligence, UAE~~~ \\
{\tt\small \{bimsara.pathiraja,malitha.gunawardhana,muhammad.haris\}@mbzuai.ac.ae}
}


\maketitle

\begin{abstract}
Albeit achieving high predictive accuracy across many challenging computer vision problems, recent studies suggest that deep neural networks (DNNs) tend to make overconfident predictions, rendering them poorly calibrated. Most of the existing attempts for improving DNN calibration are limited to classification tasks and restricted to calibrating in-domain predictions. Surprisingly, very little to no attempts have been made in studying the calibration of object detection methods, which occupy a pivotal space in vision-based security-sensitive, and safety-critical applications. In this paper, we propose a new train-time technique for calibrating modern object detection methods. It is capable of jointly calibrating multiclass confidence and box localization by leveraging their predictive uncertainties. We perform extensive experiments on several in-domain and out-of-domain detection benchmarks. Results demonstrate that our proposed train-time calibration method consistently outperforms several baselines in reducing calibration error for both in-domain and out-of-domain predictions. Our code and models are available at \url{https://github.com/bimsarapathiraja/MCCL}
\end{abstract}

\section{Introduction}
\label{section:Introduction}





\begin{figure*}
\centering
\begin{subfigure}{.32\textwidth}
    \centering
    \includegraphics[width=.95\linewidth]{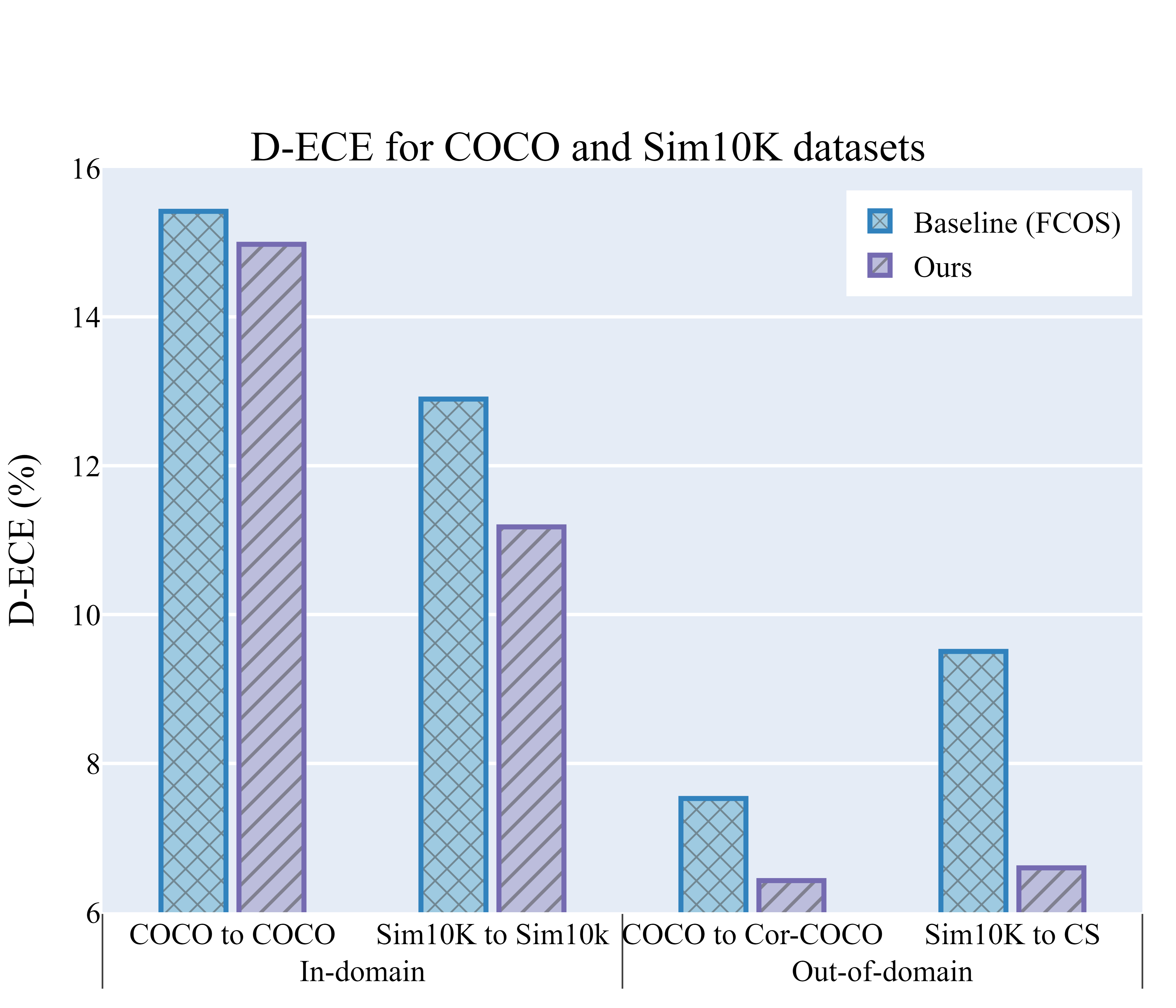}  
    \caption{}
    \label{SUBFIGURE LABEL 1}
\end{subfigure}
\begin{subfigure}{.32\textwidth}
    \centering
    \includegraphics[width=.95\linewidth]{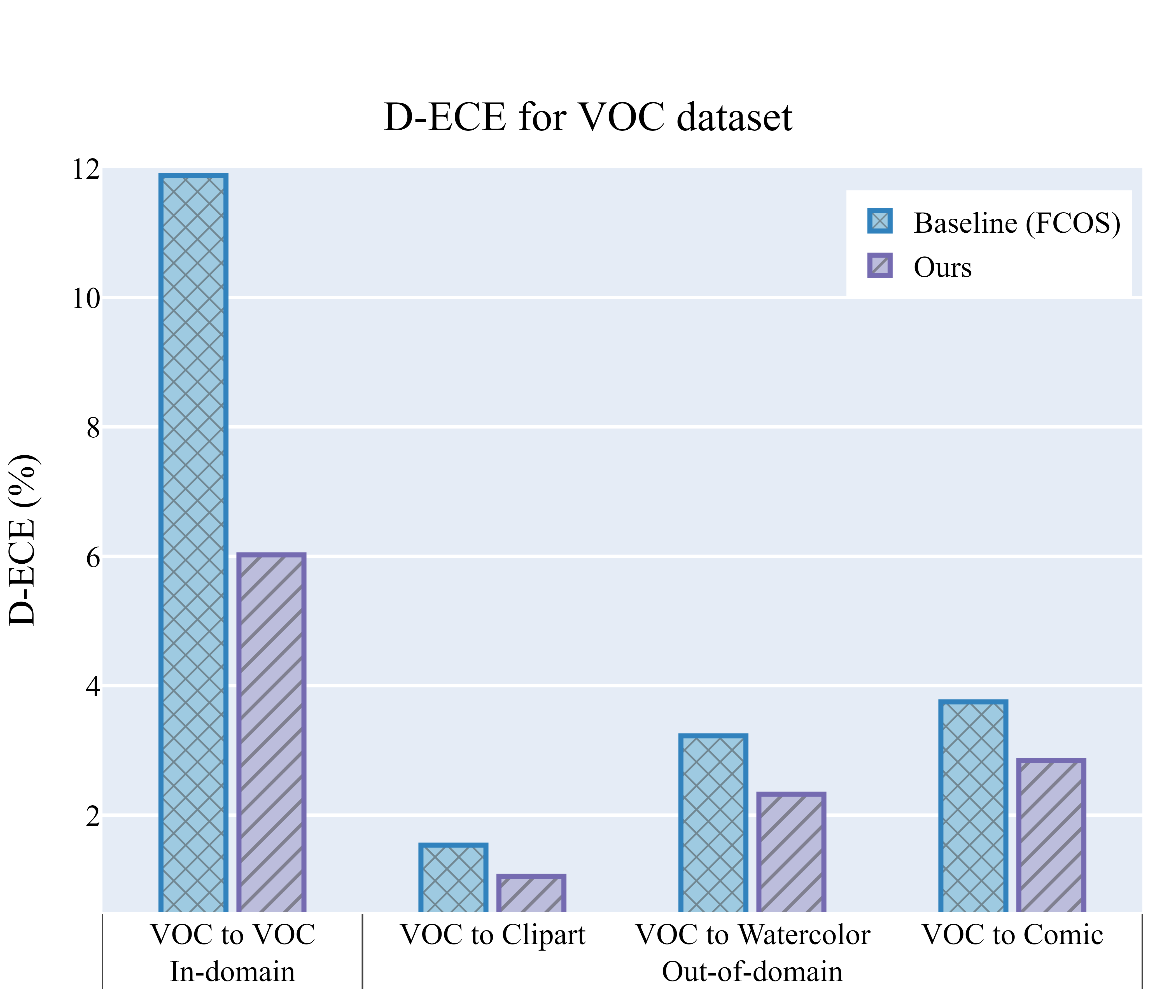}  
    \caption{}
    \label{SUBFIGURE LABEL 2}
\end{subfigure}
\begin{subfigure}{.32\textwidth}
    \centering
    \includegraphics[width=.95\linewidth]{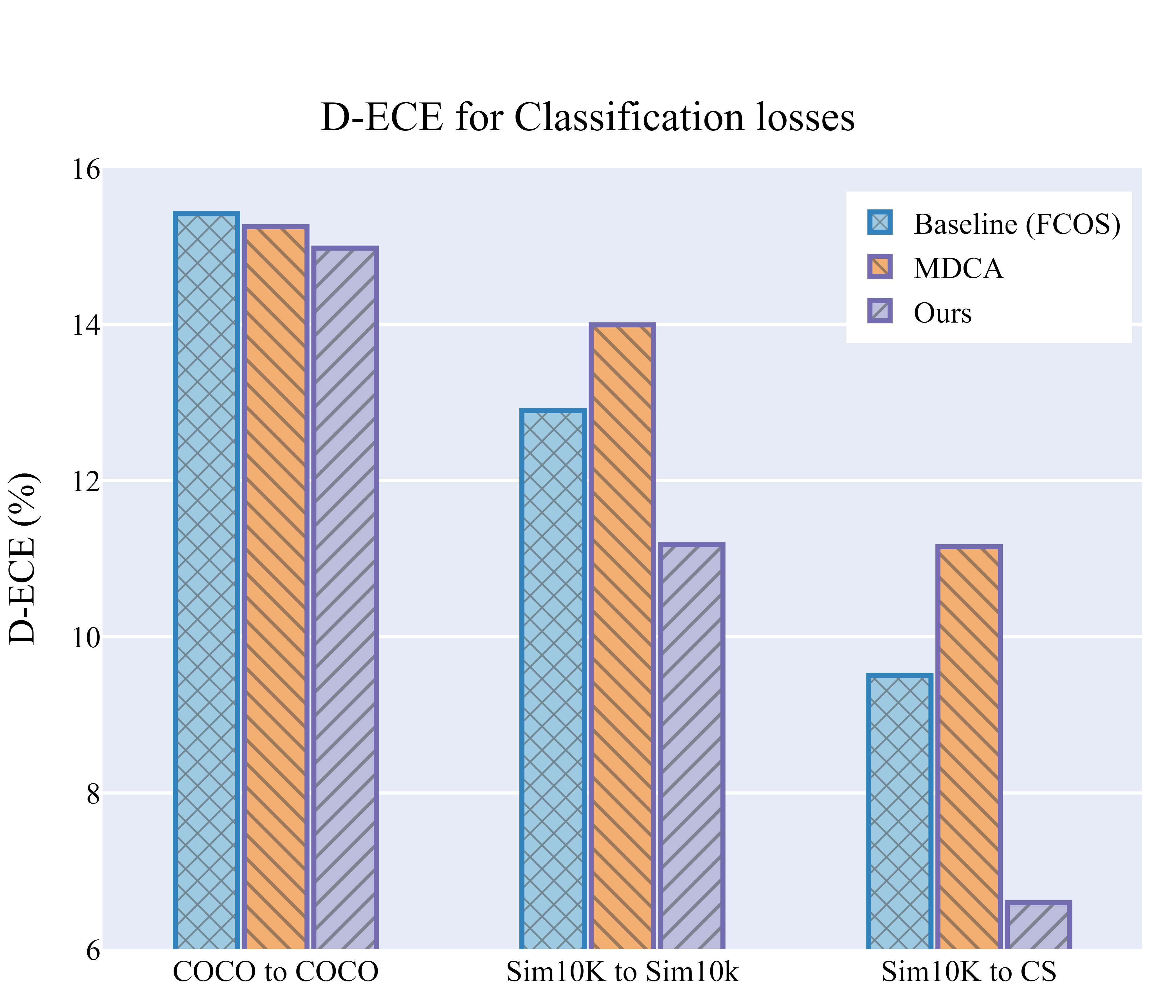}  
    \caption{}
    \label{SUBFIGURE LABEL 3}
\end{subfigure}
  \vspace{-0.5em}
\caption{DNN-based object detectors are inherently miscalibrated for both in-domain and out-of-domain predictions. Also, calibration methods for image classification are sub-optimal for object detection. Our proposed train-time calibration method for object detection is capable of reducing the calibration error (D-ECE\%) of DNN-based detectors in both in-domain and out-domain scenarios.}

\vspace*{-\baselineskip}
\label{fig:D-ECE_in_domain_out_of_domain}
\end{figure*}

Deep neural networks (DNNs) are the backbone of many top-performing systems due to their high predictive performance across several challenging domains, including computer vision \cite{he2016deep, ronneberger2015unet, mask_2017_ICCV, tian2019fcos, zhu2021deformable} and natural language processing \cite{devlin2018bert, brown2020language}. However, some recent works \cite{guo2017calibration, ovadia2019can, wenzel2020hyperparameter,havasi2020training} report that DNNs are susceptible to making overconfident predictions, which leaves them miscalibrated. This not only spurs a mistrust in their predictions, but more importantly, could lead to disastrous consequences in several safety-critical applications, such as healthcare diagnosis \cite{dusenberry2020analyzing, sharma2017crowdsourcing}, self-driving cars \cite{grigorescu2020survey}, and legal research tools \cite{yu2019s}. For instance, in self-driving cars, if the perception component wrongly detects a stop sign as a speed limit sign with high confidence, it can potentially lead to disastrous outcomes. 


Several strategies have been proposed in the recent past for improving model calibration. A simple calibration technique is a post-processing step that re-scales the outputs of a trained model using parameters which are learnt on a hold-out portion of the training set \cite{guo2017calibration}. Despite being easy to implement, these post-processing approaches are restrictive. They assume the availability of a hold-out set, which is not always possible in many real-world settings. Another route to reducing calibration error is train-time calibration techniques, which intervene at the training time by involving all model parameters. Typically train-time calibration methods feature an auxiliary loss term that is added to the application-specific loss function to regularize predictions \cite{kumar2018trainable, mukhoti2020calibrating, hebbalaguppe2022stitch, Liu_2022_CVPR}. 

We note that almost all prior efforts towards improving model calibration target the task of visual image classification. Surprisingly, little to no noticeable attempts have been made in studying the calibration of visual object detection models. Visual object detection methods account for a major and critical part of many vision-based decision-making systems. Moreover, most of the current calibration techniques only aim at reducing calibration error for in-domain predictions. However, in many realistic settings, it is likely that, after model deployment, the incoming data distribution could continuously change from the training data distribution. In essence, the model should be well-calibrated for both in-domain and out-of-domain predictions.

To this end, in this paper, we aim to study the calibration of (modern) deep learning-based object detection methods. In this pursuit, we observe that, (a) object detection methods are intrinsically miscalibrated, (b) besides displaying noticeable calibration errors for in-domain predictions, they are also poorly calibrated for out-of-domain predictions and, (c) finally, the current calibration techniques for classification are sub-optimal for object detection (\Cref{fig:D-ECE_in_domain_out_of_domain}). Towards improving the calibration performance of object detection methods, inspired by the train-time calibration route, we propose a new train-time calibration approach aims at jointly calibrating the predictive multiclass confidence and bounding box localization. 

\noindent\textbf{Contributions:} \textbf{(1)} We study the relatively unexplored direction of calibrating modern object detectors and observe that they are intrinsically miscalibrated in both in-domain and out-of-domain predictions. Also, the existing calibration techniques for classification are sub-optimal for calibrating object detectors. \textbf{(2)} We propose a new train-time calibration method for detection, at the core of which is an auxiliary loss term, which attempts to jointly calibrate multiclass confidences and bounding box localization. We leverage predictive uncertainty in multiclass confidences and bounding box localization. \textbf{(3)} Our auxiliary loss term is differentiable, operates on minibatches, and can be utilized with other task-specific loss functions. \textbf{(4)} We perform extensive experiments on challenging datasets, featuring several in-domain and out-of-domain scenarios. Our train-time calibration method consistently reduces the calibration error across DNN-based object detection paradigms, including FCOS \cite{tian2019fcos} and Deformable DETR \cite{zhu2021deformable}, both in in-domain and out-of-domain predictions.



\section{Related works}

\noindent\textbf{Post-processing calibration methods:} A simple approach to calibration is a post-processing step, which re-scales the outputs of a trained model using some parameters that are learned on the hold-out portion of the training set. Temperature scaling (TS), which is an adaptation of Platt scaling \cite{platt1999probabilistic}, is a prominent example. It divides the logits (pre-softmax activations) from a trained network with a fixed temperature parameter ($\mathrm{T}>0$) that is learned using a hold-out validation set. An obvious limitation of TS is that it decreases the confidence of the whole (confidence) vector, including the confidence of the correct class. Beyond using a single temperature parameter ($\mathrm{T}$), some works uses a matrix ($\mathrm{M}$) to to transform the logits. The matrix ($\mathrm{M}$) is also learnt using a hold-out validation set. Dirichlet calibration (DC) employed Dirichlet distributions to generalize the Beta-calibration \cite{kull2017beta} method, originally proposed for binary classification, to a multi-class setting. DC is realized as an extra layer in a neural network whom input is log-transformed class probabilities. The work of \cite{bohdal2021meta} proposed a differentiable approximation of expected calibration error (ECE) and utilizes it in a meta-learning framework to obtain well-calibrated models. Islam et al. \cite{islam2021class} achieved class-distribution-aware calibration using temperature scaling (TS) and label smoothing (LS) \cite{szegedy2016rethinking} for long-tailed visual recognition. Majority of the aforementioned work address in-domain calibration. Recently, \cite{tomani2021post} proposed to gradually perturb the hold-out validation set for simulating out-of-domain prior to learning the temperature parameter ($\mathrm{T}$). Despite being easy-to-implement and effective, TS methods require a hold-out validation set, which is not readily available in many realistic scenarios.

\noindent\textbf{Train-time calibration techniques:} Another approach to improving model calibration are train-time calibration techniques. Brier score is considered one of the earliest attempts for calibrating binary probabilistic forecast \cite{brier1950verification}. Some recent works report that models trained with negative log-likelihood (NLL) are prone to making overconfident predictions. A dominant class in train-time methods typically propose an auxiliary loss term that is used in conjunction with NLL. For instance, \cite{pereyra2017regularizing} utilized the Shanon entropy to penalize overconfident predictions. Similarly, Muller et al. \cite{muller2019does} showed that label smoothing \cite{szegedy2016rethinking} also improves calibration. Recently, \cite{Liu_2022_CVPR} introduced a margin into the label smoothing technique to obtain well-calibrated models. While re-visiting focal loss (FL) \cite{lin2018focal}, \cite{mukhoti2020calibrating} demonstrated that it is capable of implicitly calibrating DNNs.
Liang et al. \cite{liang2020imporved} incorporated the difference between confidence and accuracy (DCA) as an auxiliary loss term with the Cross-Entropy loss to achieve model calibration. Likewise, \cite{kumar2018trainable} developed MMCE loss for calibrating DNNs, which is formulated using a reproducible kernel in Hilbert space \cite{gretton2013introduction}. Most of these methods only calibrate the confidence of the predicted label ignoring the confidences of non-predicted classes. Recently, \cite{hebbalaguppe2022stitch} proposed an auxiliary loss term for calibrating the whole confidence vector.

\noindent\textbf{Probabilistic and non-probabilistic methods:}
Many probabilistic approaches stem from Bayesian formalism \cite{bernardo2009bayesian}, which assumes a prior distribution over the neural network (NN) parameters, and training data is leveraged to obtain the posterior distribution over the NN parameters. This posterior is then used to estimate the predictive uncertainty. 
The exact Bayesian inference is computationally intractable, consequently, we can see approximate inference methods, including variational inference \cite{blundell2015weight,louizos2016structured}, and stochastic expectation propagation \cite{hernandez2015probabilistic}. 
A non-probabilistic approach is ensemble learning that can be used to quantify uncertainty; it uses the empirical variance of the network predictions. 
Ensembles can be created with the differences in model hyperparameters \cite{wenzel2020hyperparameter}, random initialization of weights and random shuffling of training data \cite{lakshminarayanan2017simple}, dataset shift \cite{ovadia2019can}, and Monte Carlo (MC) dropout \cite{gal2016dropout,zhang2019confidence}. 
In this work, we propose to use MC dropout \cite{gal2016dropout} to quantify predictive uncertainty both in class confidences and the bounding localization. It allows creating a distribution over both outputs from a typical DNN-based object detector. The naive implementation of MC dropout can incur high computational cost for large datasets and network architectures during model training. So, we resort to an efficient implementation of MC dropout that greatly reduces this computational overhead.

We note that, almost all prior work for addressing calibration is targeted at classification task \cite{guo2017calibration,kull2017beta,kumar2018trainable, hebbalaguppe2022stitch,Liu_2022_CVPR}, and no noticeable study has been published that strives to improve the calibration of object detection methods, especially for out-of-domain predictions. In this paper, we explore the problem of calibrating object detectors and observe that they are inherently miscalibrated for both in-domain and out-domain predictions. To this end, we propose a train-time calibration method aimed at jointly calibrating multiclass confidence and bounding box localization.

\section{Method}

\subsection{Defining and Measuring Calibration}

\noindent\textbf{Calibration for classification:} A perfectly calibrated model for (image) classification outputs class confidences that match with the predictive accuracy. If the accuracy is less than the confidence, then the model is overconfident and if the accuracy is higher than the confidence, then the model is underconfident. 
Let $\mathcal{D}$ = $\left<(\mathbf{x}_{i},y_{i}^{*})\right>_{i=1}^{N}$ denote a dataset consisting of N examples drawn from a joint distribution $\mathcal{D(X, Y)}$, 
where $\mathcal{X}$ is an input space and $\mathcal{Y}$ is the label space. For each sample $\mathbf{x}_{i}$ $\in$  $\mathcal{X}$, $y_{i}^{*}$  $\in$  $\mathcal{Y}$ = $\left\{1,2,...K \right\}$ is the corresponding ground truth class label.
Let $\mathbf{s}$ $\in$  $\mathbb{R}^K$ be the vector containing the predicted confidences of all $K$ classes, and $\mathbf{s}_{i}[y]$ be the confidence predicted for a class $y$ on a given input example $\mathbf{x}_{i}$. The model is said to be perfectly calibrated when, for each sample $(\mathbf{x},y)$ $\in$ $\mathcal{D}$:

\vspace{-1em}
\begin{equation}
   \mathbb{P}(y=y^* | \mathbf{s}[y] = s) = s
   \label{eq:calibration 2}
\end{equation}

where $\mathbb{P}(y=y^* | s[y] = s)$ is the accuracy for each confidence scores in $\mathbf{s}$.

\noindent\textbf{Calibration for object detection:} Contrary to classification, in object detection, the dataset contains the ground-truth annotations for each object in an image, specifically the object localization information and the associated object categories. Let $\mathbf{b}^{*} \in \mathcal{B} = [0,1]^4$ be the bounding box annotation of the object and $y^{*}$ be the corresponding class label. 
%
%
The prediction from an object detection model consists of a class label $\hat{y}$, with a confidence score $\hat{s}$ and a bounding box $\hat{\mathbf{b}}$. Unlike classification, for object detection, precision is used instead of accuracy for calibration. Therefore, an object detector is perfectly calibrated when \cite{Kueppers_2020_CVPR_Workshops}: 
\vspace{-0.5em}
\begin{equation}
\mathbb{P}(m=1 | \hat{s} =s, \hat{y} = y, \hat{\mathbf{b}}= \mathbf{b})  =  s
  \label{eq:calibration2}
\end{equation}

\centerline{$\forall s \in [0,1], y \in \mathcal{Y}, \mathbf{b} \in [0,1]^4 $}  

where $m=1$ denotes a correctly classified prediction i.e. whose $\hat{y}$ matches with the $y^{*}$ and the Intersection-over-Union (IoU) between $\hat{\mathbf{b}}$ and $\mathbf{b}^{*}$ is greater than a certain threshold $\gamma$. Thus, $\mathbb{P}(m=1)$ amounts to approximating $\mathbb{P} (\hat{y}=y^{*}, \hat{\mathbf{b}} = \mathbf{b}^{*})$ with a certain IoU threshold $\gamma$.


\noindent\textbf{Measuring miscalibration for classification and object detection:} For classification, the expected calibration error (ECE) is used to measure the miscalibration of a model. The ECE measures the expected deviation of the predictive accuracy from the estimated confidence \cite{guo2017calibration,naeini2015obtaining, Kueppers_2020_CVPR_Workshops}:
\vspace{-0.5em}
\begin{equation}
    \label{eq:ECE1}
    \mathbb{E}_{\hat{s}} \left [ \left| \mathbb{P}(\hat{y}=y| \hat{s}=s)-s\right| \right]
\end{equation}

As $\hat{s}$ is a continuous random variable, the ECE is approximated by binning the confidence space of $\hat{s}$ into $N$ equally spaced bins. Therefore, ECE is approximated by \cite{naeini2015obtaining}:
\vspace{-0.5em}
\begin{equation}
    \label{eq:ECE-approximated}
    \mathrm{ECE} = \sum_{n=1}^{N} \frac{|I(n)|}{\mathcal{|D|} } . \left| \mathrm{acc}(n) -\mathrm{conf}(n)\right| 
\end{equation}

where $|I(n)|$ is the number of examples in the $n^{th}$ bin, and ${\mathcal{|D|}}$ is the total number of examples. $\mathrm{acc}(n)$ and $\mathrm{conf}(n)$ denote the average accuracy and average confidence in the $n^{th}$ bin, respectively. Although the ECE measure can be used for measuring miscalibration of object detectors, it fails to reflect the calibration improvement when additional box coordinates are used for calibration since the ECE considers confidence of each example independent of the box properties to apply binning and to calculate an average precision. In this work, we use location-dependent calibration, termed as detection ECE (D-ECE). It is defined as the expected deviation of the observed precision with respect to the given box properties. 
\vspace{-0.5em}
\begin{equation}
    \label{eq:DECE}
    \mathbf{E}_{\hat{s} \hat{\mathbf{b}}} \left[  \left | \mathbb{P}(m = 1| \hat{s} = s, \hat{y} = y , \hat{\mathbf{b}}= \mathbf{b}) -s         \right |         \right]
\end{equation}

Similar to ECE, the multidimensional D-ECE is calculated by partitioning both the confidence and box property spaces in each dimension $k$ into $N_k$ equally spaced bins. Thus, D-ECE is given by \cite{Kueppers_2020_CVPR_Workshops}:
\vspace{-0.5em}
\begin{equation}
    \mathrm{\operatorname{D-ECE}}_{k} = \sum_{n=1}^{N_{total}} \frac{|I(n)|}{\mathcal{|D|} } . \left|\mathrm{prec}(n) - \mathrm{conf}(n)\right|
    \label{eq:d_ece_metric}
\end{equation}

where $N_{total}$ is the total number of bins. $\mathrm{prec}(n)$ and $\mathrm{conf}(n)$ denote the average precision and confidence in each bin, respectively. 

\subsection{Proposed train-time calibration: MCCL}

This section describes our new train-time calibration method at the core of which is an auxiliary loss function. This auxiliary loss formulation aims at jointly calibrating the multiclass confidence and bounding box localization. It is based on the fact that, the modern object detectors (based on DNNs) predict a confidence vector along with the bounding box parameters. 
The two key quantities to our loss function are (1) the predictive certainty in class logits and the bounding box localization and, (2) the class-wise confidence after computing class-wise logits mean (termed mean logits based class-wise confidence hereafter) and mean bounding box localization. 
The predictive certainty in class-wise logits is used in-tandem with the mean logits based class-wise confidence to calibrate the multi-class confidence scores. While, the predictive certainty in the bounding box prediction is used to calibrate the bounding box localization. Instead of inputting the class-wise logits and predicted bounding box parameters to the classification loss and regression loss in task-specific detection losses, we input the class-wise mean logits and mean bounding box parameters, respectively. We first describe how to compute the mean logits based class-wise confidence, mean bounding box parameters, and the certainty in both class logits and bounding box localization.


%


\noindent\textbf{Quantifying means and certainties:} For the $n^{th}$ positive location, we aim to quantify the mean logits based class-wise confidence $\bar{\mathbf{s}}_{n} \in \mathbb{R}^{K}$ and class-wise certainty in logits $\mathbf{c}_{n} \in \mathbb{R}^{K}$ as well as the mean bounding box parameters $\bar{\mathbf{b}}_{n} \in [0,1]^J$ and certainty in bounding box localization $\mathrm{g}_{n}$. Where $J$ is the number of bounding box parameters. 
Given an input sample (image), we perform $N$ stochastic forward passes by applying the Monte-Carlo (MC) dropout \cite{gal2016dropout}. It generates a distribution over class logits and bounding box localization. Assuming one-stage object detector (e.g.,\cite{tian2019fcos}), we insert a dropout layer before the classification layer and the regression layer. Let $\mathbf{z}_{n} \in \mathbb{R}^{N \times K}$ and $\mathbf{r}_{n} \in \mathbb{R}^{N \times J}$ encode the distributions over class-wise logit scores and bounding box parameters, respectively, corresponding to $n^{th}$ positive location obtained after performing N, MC forward passes. 

We obtain the mean logits based class-wise confidence $\bar{\mathbf{s}}_{n} \in \mathbb{R}^{K}$ by first taking the mean along the first dimension of $\mathbf{z}_{n}$ to get class-wise mean logits and then applying the softmax. 
To obtain class-wise certainty $\mathbf{c}_{n}$, we first estimate the uncertainty $\mathbf{d}_{n} \in \mathbb{R}^{K}$ by computing the variance along the first dimension of $\mathbf{z}_{n}$. Then, we apply $\mathrm{tanh}$ over $\mathbf{d}_{n}$ and subtract it from 1 as: $\mathbf{c}_{n} = 1 - \mathrm{tanh}(\mathbf{d}_{n})$, where $\mathrm{tanh}$ is used to scale the uncertainty $\mathbf{d}_{n} \in [0,\inf)$ between 0 and 1. 


Similarly, we estimate the mean bounding box parameters $\bar{\mathbf{b}}_{n}$ and the certainty $\mathrm{g}_{n}$ in the bounding box parameters for the $n^{th}$ positive location. 
Let $\{\sigma^{2}_{n}\}_{j=1}^{J}$ and $\{\mu_{n}\}_{j=1}^{J}$ be the vectors ($J$ is the number of bbox parameters) comprised of variances and means of predicted bounding box parameters distribution $\mathbf{r}_{n}$. These variances and the means are computed along the first dimension of $\mathbf{r}_{n}$. We term $\{\mu_{n}\}_{j=1}^{J}$ as the mean bounding box parameters $\bar{\mathbf{b}}_{n}$.
Also, let $\mu_{n,com}$ denote the combined mean, computed as $\mu_{n,com} = \frac{1}{J} \sum_{j=1}^{J} \mu_{n,j}$. Then, we estimate the (joint) uncertainty $u_{n}$ as: 
\vspace{-0.5em}
\begin{equation}
\mathrm{u}_{n} = \frac{1}{J} \sum_{j=1}^{J} [\sigma^{2}_{n,j} + (\mu_{n,j} - \mu_{n,com})^2 ].
\end{equation}

The certainty $\mathrm{g}_{n}$ in the $n^{th}$ positive bounding box localization is then computed as: $\mathrm{g}_{n} = 1 - \mathrm{tanh}(\mathrm{u}_{n})$.





We leverage these estimated mean logits based class-wise confidence, class-wise certainty and the certainty in bounding box localization to formulate the two components of our auxiliary loss: multi-class confidence calibration (MCC), and localization calibration (LC).
For MCC, we compute the difference between the fused mean confidence and certainty with the accuracy. For LC, we calculate the deviation between the predicted bounding box overlap and the predictive certainty of the bounding box. Both quantities are computed over the mini-batch during training.

\noindent\textbf{Multi-class confidence calibration (MCC):} To achieve multi-class confidence calibration, we leverage the mean logits based class-wise confidence and class-wise certainty and fuse them by computing class-wise mean. The resulting vector is termed as the multiclass fusion of mean confidence and certainty. Then, we calculate the absolute difference between the fused vector and the accuracy as:

\vspace{-1em}
\begin{equation}
     \mathcal{L}_{MCC} =  \frac{1}{K} \sum_{k=1}^{K} \Bigg | \frac{1}{M} 
    \sum_{l=1}^{N_b} \sum_{n=1}^{N_{pos}} \mathbf{v}_{l,n}[k]  - \frac{1}{M}  \sum_{l=1}^{N_b} \sum_{n=1}^{N_{pos}} \mathbf{q}_{l,n}[k]  \Bigg |	
     \label{eq:ourloss}
\end{equation}

where $M=N_b \times N_{pos}$. $N_b$ is the number of samples in the minibatch and $N_{pos}$ represents the number of positive locations. 
%
$\mathbf{q}_{l,n}[k]$ = 1 if $k$ is the ground truth class of the bounding box predicted for the $n^{th}$ location in the $l^{th}$ sample. $\mathbf{v}_{l,n}[k] = (\bar{\mathbf{s}}_{l,n}[k] +  \mathbf{c}_{l,n}[k])/2$, where $\bar{\mathbf{s}}_{l, n}[k]$ and $\mathbf{c}_{l, n}[k]$ are the mean confidence and the certainty, respectively, for the class $k$ of the $n^{th}$ positive location in the $l^{th}$ sample. 
The $\mathcal{L}_{MCC}$ is capable of calibrating the confidence of both the predicted label and non-predicted labels. It penalizes the model if, for a given class $k$, the fusion (of mean logits based class-wise confidence and certainty in class-wise logits) across minibatch deviates from the average occurrence of this class across minibatch. 


\noindent\textbf{Localization calibration (LC):} We calibrate the localization component by leveraging the certainty in bounding box prediction. Next, we compute the absolute difference between the mean bounding box overlap (with the ground truth) and the certainty in the bounding box prediction: 
\vspace{-1.1em}
\begin{equation}
    \mathcal{L}_{LC} = \frac{1}{N_b} \sum_{l=1}^{N_b} \frac{1}{N_{pos}^{l}} \sum_{n=1}^{N_{pos}^{l}}  
\left | \left [\mathrm{IoU}(\bar{\mathbf{b}}_{n,l}, \mathbf{b}_{n,l}^{*}) - \mathrm{g}_{n,l}  \right] \right|   
\end{equation}


 where $N_{pos}^{l}$ denotes the number of positive bounding box regions in the $l^{th}$ sample. $\bar{\mathbf{b}}_{n,l}$ denotes the mean bounding box parameters and $\mathrm{g}_{n,l}$ is the certainty for the $n^{th}$ positive bounding box prediction from $l^{th}$ sample. 

\noindent Both $\mathcal{L}_{MCC}$ and $\mathcal{L}_{LC}$ operate over the mini-batches, and we combine them to get our new auxiliary loss term $\mathcal{L}_{MCCL-aux}$:
\vspace{-0.5em}
\begin{equation}
\mathcal{L}_{MCCL-aux} = \mathcal{L}_{MCC} + \beta \mathcal{L}_{LC}
\label{eq:joint_loss}
\end{equation}

where $\beta$ is a hyperparameter to control the relative contribution of $\mathcal{L}_{LC}$ to the overall loss $\mathcal{L}_{MCCL-aux}$. 



\begin{table*}[!htp]
\centering
\scalebox{0.86}{
\tabcolsep=0.2cm
\begin{tabular}{lllllllllll}
  \toprule
  \multicolumn{11}{c}{In-domain performance}\\
  \midrule
Methods  &  \multicolumn{2}{c}{Sim10K} & \multicolumn{2}{c}{KITTI}  & \multicolumn{2}{c}{CS}    & \multicolumn{2}{c}{COCO}  &\multicolumn{2}{c}{VOC}\\
        & D-ECE     & AP@0.5           & D-ECE  & AP@0.5             & D-ECE  & AP@0.5          & D-ECE   & AP@0.5       & D-ECE  & mAP\\
    
    \midrule
    \multirow{2}{4em}{Baseline (FCOS)}
    & \multirow{2}{3em}{12.90} 
    & \multirow{2}{3em}{87.45}
    & \multirow{2}{3em}{9.54}
    & \multirow{2}{3em}{94.54}
    & \multirow{2}{3em}{9.40}
    & \multirow{2}{3em}{70.48}
    & \multirow{2}{3em}{15.42}
    & \multirow{2}{3em}{54.91}
     & \multirow{2}{3em}{11.88}
      & \multirow{2}{3em}{59.68}\\
     &   &   &  &  &  & \\

    Ours (MCCL)   & \textbf{11.18}  & 86.47  & 
    \textbf{7.79} & 93.76 & 
    \textbf{7.64} & 70.22 & 
    \textbf{14.94}  & 54.85 &
    \textbf{6.02}  & 59.17\\
  
  \bottomrule
\end{tabular}}
\vspace{-0.5em}
\caption{In-domain calibration performance (in D-ECE\%) on five challenging datasets, including Sim10K, KITTI, Cityscapes (CS), COCO and VOC. Best results are in bold.}
\label{tab:in_domain_perfor}
\end{table*}

\begin{table*}[!htp]
\centering
\scalebox{0.86}{
\tabcolsep=0.2cm
\begin{tabular}{lllllllllll}
  \toprule
  \multicolumn{11}{c}{Out-of-domain performance}\\
  \midrule
  Methods    &  \multicolumn{2}{c}{Sim10K$\,\to\,$CS}  & \multicolumn{2}{c}{KITTI$\,\to\,$CS}  & \multicolumn{2}{c}{CS$\,\to\,$CS-F} &
  \multicolumn{2}{c}{COCO$\,\to\,$Cor-COCO} &
  \multicolumn{2}{c}{CS$\,\to\,$BDD100K} \\
  
       & D-ECE  & AP@0.5  & D-ECE  & AP@0.5 & D-ECE  & AP@0.5 & D-ECE  & AP@0.5 &D-ECE  & AP@0.5\\
  
  \midrule
    \multirow{2}{4em}{Baseline (FCOS)}
    & \multirow{2}{3em}{9.51} 
    & \multirow{2}{3em}{45.18}
    & \multirow{2}{3em}{7.53}
    & \multirow{2}{3em}{38.11}
    & \multirow{2}{3em}{11.18}
    & \multirow{2}{3em}{19.81}
    & \multirow{2}{3em}{15.90}
    & \multirow{2}{3em}{30.01}
    & \multirow{2}{3em}{18.82}
     & \multirow{2}{3em}{14.18} \\
     &   &   &  &  &  & \\
    
     Ours (MCCL)    & \textbf{6.60 } & 44.30  
     & \textbf{6.43}  & 38.73  
     & \textbf{8.97}  & 19.54
     &  \textbf{14.45} &29.96 
     &  \textbf{16.12}  & 14.20\\

  \bottomrule
  
\end{tabular}}
\vspace{-0.5em}
\caption{Out-of-domain calibration performance (in D-ECE\%) on five challenging domain shifts.}
\label{tab:out of domain}
\end{table*}

\begin{table*}[!htp]
\centering
\tabcolsep=0.425cm
\scalebox{0.86}{
\begin{tabular}{lllllll}
  \toprule
  \multicolumn{7}{c}{Out-of-domain performance}\\
  \midrule
  Methods   & 
  \multicolumn{2}{c}{VOC$\,\to\,$clipart}  & 
  \multicolumn{2}{c}{VOC$\,\to\,$ watercolor}  & \multicolumn{2}{c}{VOC$\,\to\,$comic}  \\
 
        & D-ECE  & mAP & D-ECE  & mAP & D-ECE  & mAP\\
  
  \midrule
    Baseline (FCOS)   & 1.54  & 14.57  & 3.23  & 24.23 & 3.75  & 9.89\\
     Ours (MCCL)     & \textbf{1.06}  & 13.71  
     & \textbf{2.33}  & 28.70 
     & \textbf{2.84}  & 11.50\\
  
  \bottomrule
  
\end{tabular}}
\vspace{-0.5em}
\caption{\label{tab:domain-voc}Out-of-domain calibration performance on three challenging domain drifts.}
\end{table*}

\section{Experiments}

\label{sec:datasets}

\noindent\textbf{Datasets:} To evaluate the in-domain calibration performance, we use the following five datasets: Sim10K ~\cite{johnson2017driving}, KITTI~\cite{geiger2012we}, Cityscapes (CS)~\cite{Cordts2016Cityscapes}, COCO~\cite{cocodataset}, and PASCAL VOC(2012)~\cite{Everingham15}. 
\textbf{Sim10K} \cite{johnson2017driving} contains synthetic images of the car category, and offers 10K images which are split into 8K for training, 1K for validation and 1K for testing.
\textbf{Cityscapes} \cite{Cordts2016Cityscapes} is an urban driving scene dataset and consists of 8 object categories. 
It has 2975 training images and 500 validation images, which are used for evaluation. 
\textbf{KITTI} \cite{geiger2012we} is similar to Cityscapes as it contains images of road scenes with a wide view of the area, except that KITTI images were captured with a different camera setup. Following prior works, we consider car class for experiments.
We use train2017 version of \textbf{MS-COCO}~\cite{cocodataset} and it offers 118K training images, 5K validation images, and 41K test images. 
\textbf{PASCAL VOC 2012}~\cite{Everingham15} consists of 5,717 training and 5,823 validation images, and provides  bounding box annotations for 20 classes.
For evaluating out-of-domain calibration performance, we use Sim10K to CS, KITTI to CS, CS to Foggy-CS, COCO to Cor-COCO, CS to BDD100K\cite{yu2018bdd100k}, VOC to Clipart1k\cite{inoue_2018_cvpr}, VOC to Watercolor2k\cite{inoue_2018_cvpr}, and VOC to Comic2k\cite{inoue_2018_cvpr}.
\textbf{Foggy Cityscapes} (CS-F)\cite{sakaridis2018semantic} dataset is developed using Cityscapes dataset \cite{Cordts2016Cityscapes} by simulating foggy weather leveraging the depth maps in Cityscapes with three levels of foggy weather.
\textbf{Cor-COCO} is a corrupted version of MS-COCO val2017 dataset for out-of-domain evaluation, and is constructed by introducing random corruptions with severity levels defined in \cite{hendrycks2019benchmarking}.
\textbf{Clipart1k} \cite{inoue_2018_cvpr} contains 1K images, which are split into 800 for training and 200 for validation, and shares 20 object categories with PASCAL VOC.
Both \textbf{Comic2k}\cite{inoue_2018_cvpr} and \textbf{Watercolor2k}\cite{inoue_2018_cvpr} are comprised of 1K training images and 1K test images, and share 6 categories with Pascal VOC. 
\textbf{BDD100k} \cite{yu2020bdd100k} offers 70K training images, 20K test images and 10K validation images. We use validation set for out-of-domain evaluation.

\noindent\textbf{Implementation Details:} For all experiments, we use Tesla V100 GPUs. 
For COCO experiments, we use 8 GPUs and follow training configurations reported in \cite{tian2019fcos}. For experiments on all other datasets, we utilize 4 GPUs and follow training configurations listed in \cite{hsu2020every}. We chose $\beta$ in \Cref{eq:joint_loss} from $\{0.01, 1\}$. For further training details, we refer to the supplementary material.



\noindent\textbf{Evaluation metrics:} We use D-ECE metric defined in \Cref{eq:d_ece_metric} at IoU of 0.5 to measure calibration performance. Note that, in addition to classification scores, it takes into account the calibration of center-x, center-y, width, and height of the predicted box. For reporting detection performance, we use mAP and AP@0.5 metrics.



\noindent\textbf{Baselines:} We evaluate our train-time calibration method against models trained with task-specific losses of a CNN-based object detector, namely FCOS \cite{tian2019fcos}, and ViT-based object detector, namely Deformable DETR\cite{zhu2021deformable}. We then compare with the temperature scaling post-hoc method and further with the recently proposed auxiliary loss functions for classification, including MDCA\cite{hebbalaguppe2022stitch} and AvUC\cite{krishnan2020improving}.




\subsection{Results}
\label{subsection:Results}

\begin{figure*}[h!]
\centering
\begin{subfigure}{.24\textwidth}
    \centering
    \includegraphics[width=.95\linewidth, height=1.8cm]{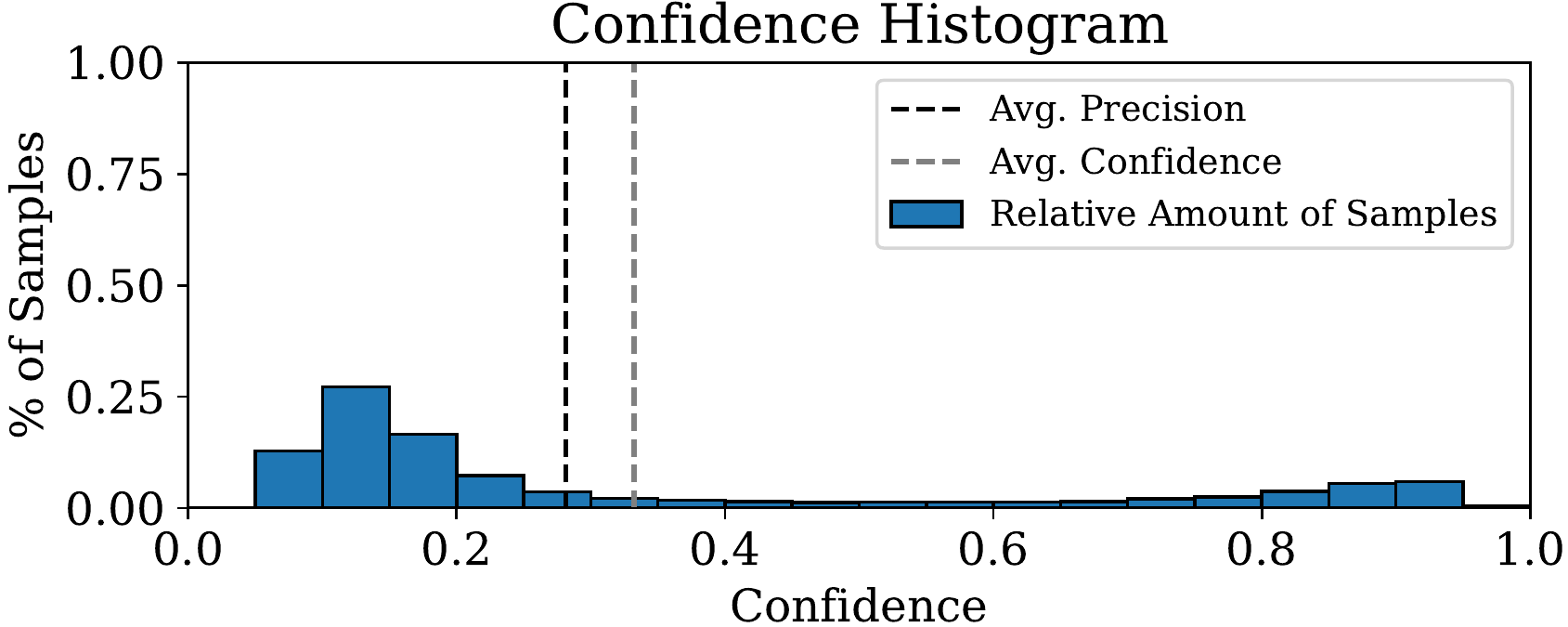}  
    \caption{Sim10K - baseline (FCOS)}
    \label{fig:sim10K_base}
\end{subfigure}
\begin{subfigure}{.24\textwidth}
    \centering
    \includegraphics[width=.95\linewidth, height=1.8cm]{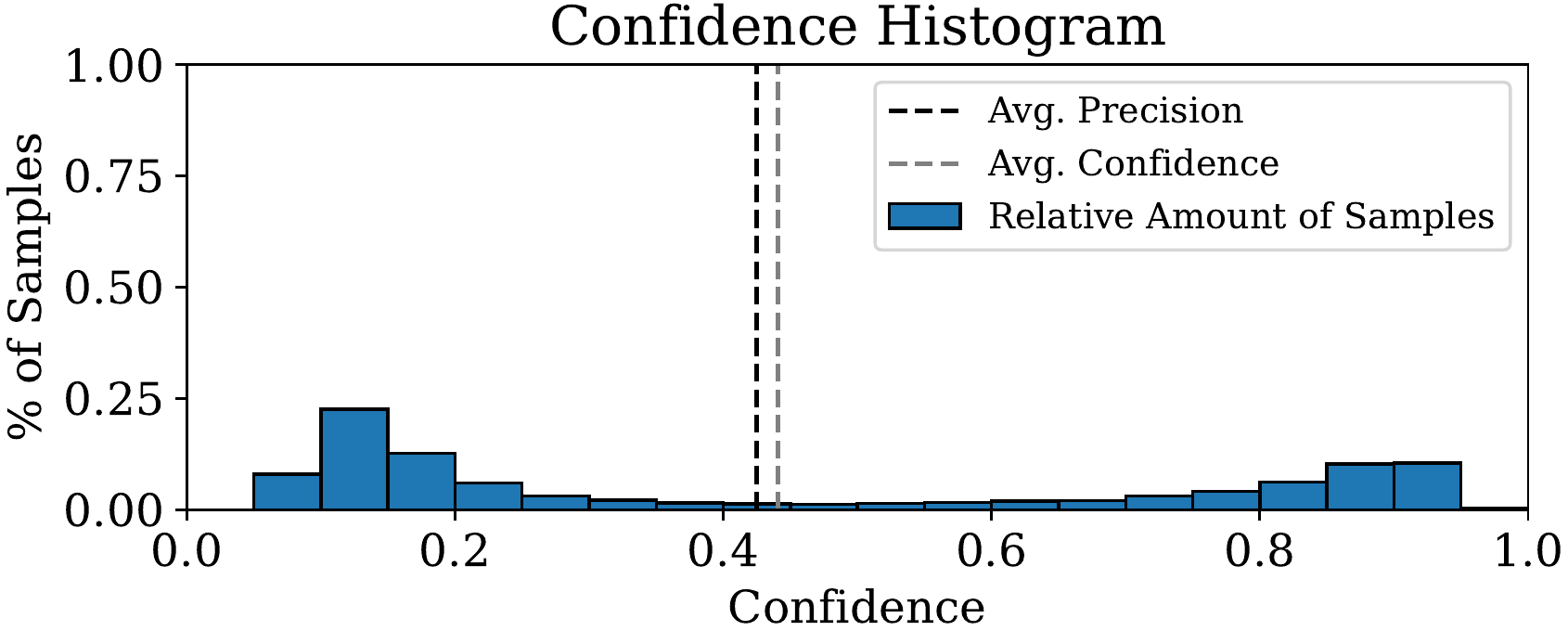}  
    \caption{Sim10K - Ours}
    \label{fig:sim10K_ours}
\end{subfigure}
\begin{subfigure}{.24\textwidth}
    \centering
    \includegraphics[width=.95\linewidth, height=1.8cm]{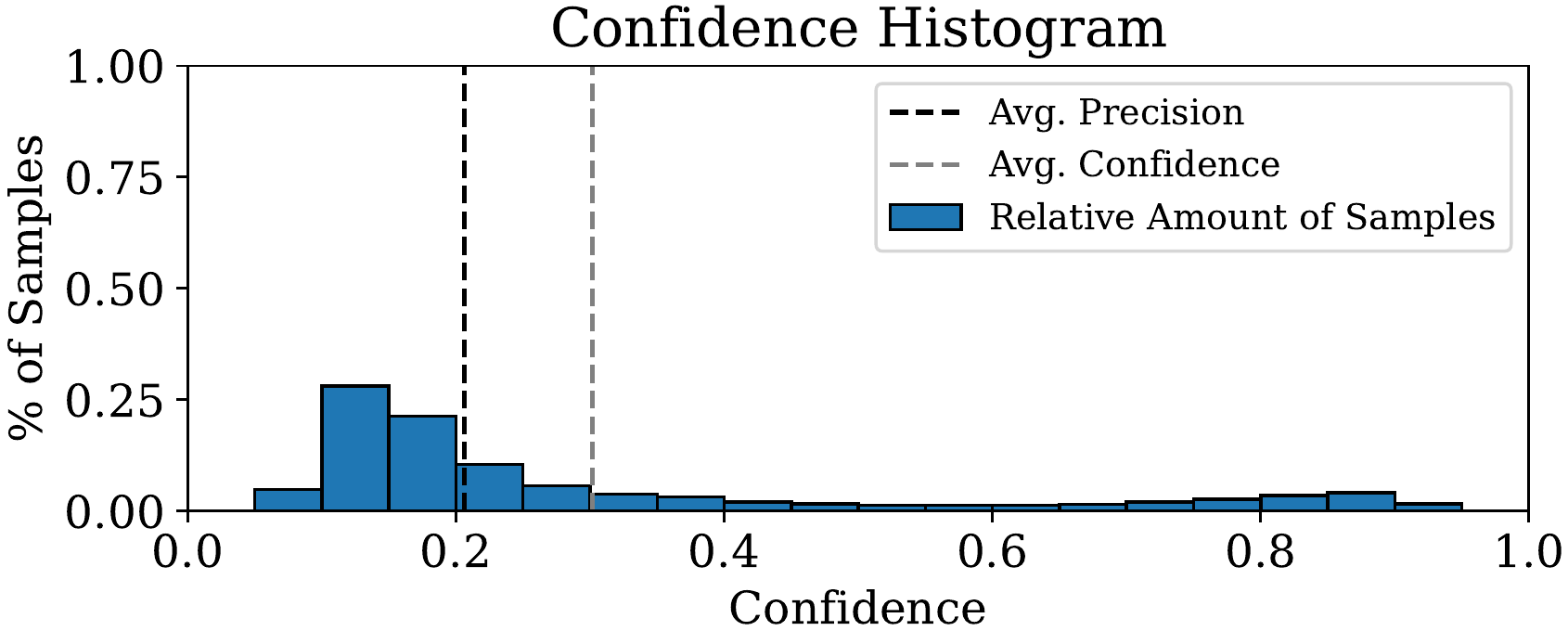}  
    \caption{CS to CS-F - baseline (FCOS)}
    \label{fig:CS_FCS_base}
\end{subfigure}
\begin{subfigure}{.24\textwidth}
    \centering
    \includegraphics[width=.95\linewidth, height=1.8cm]{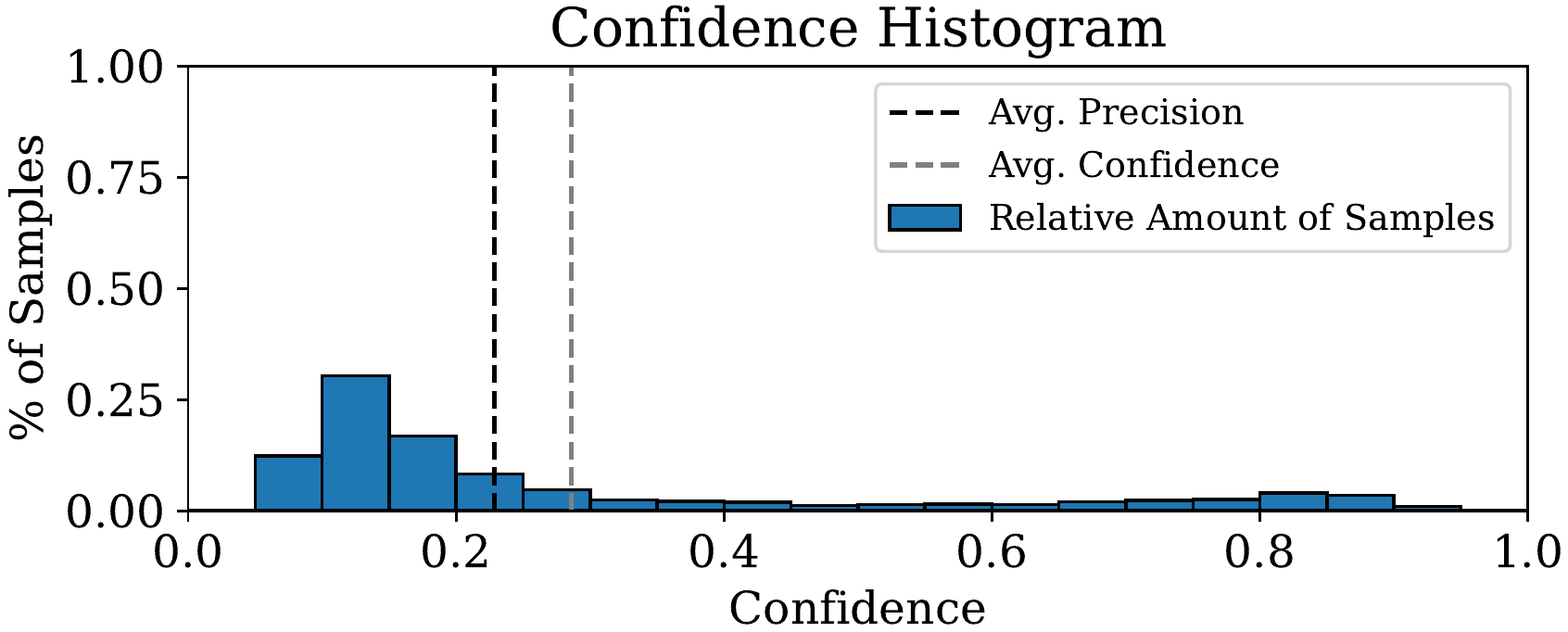}  
    \caption{CS to CS-F - Ours}
    \label{fig:CS_FCS_ours}
\end{subfigure}
\vspace{-0.5em}
\caption{Confidence histograms for baseline and our method.}
\vspace{-0.5em}
\label{fig:confidence_histogram}
\end{figure*}

\begin{figure*}[!htp]
\centering
\begin{subfigure}{.24\textwidth}
    \centering
    \includegraphics[width=.95\linewidth, height=1.8cm]{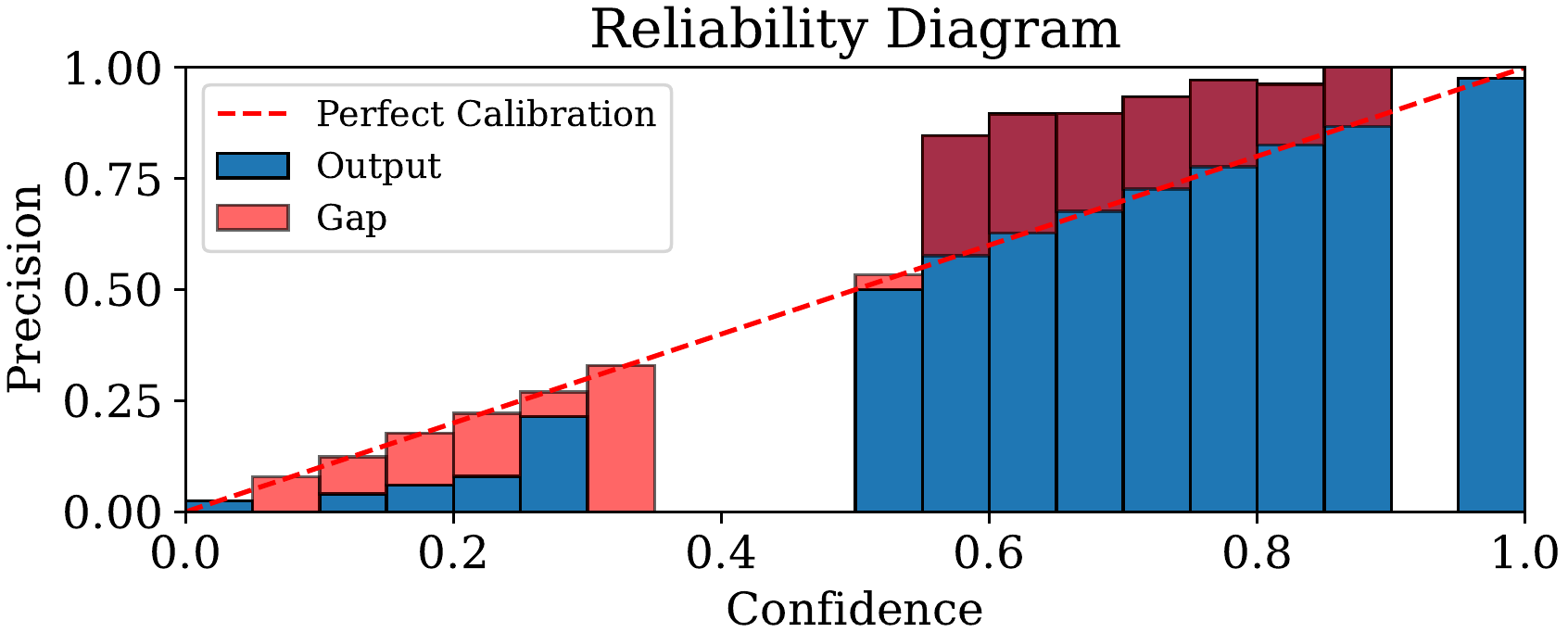}  
    \caption{VOC - baseline (FCOS)}
    \label{fig:VOC_base}
\end{subfigure}
\begin{subfigure}{.24\textwidth}
    \centering
    \includegraphics[width=.95\linewidth, height=1.8cm]{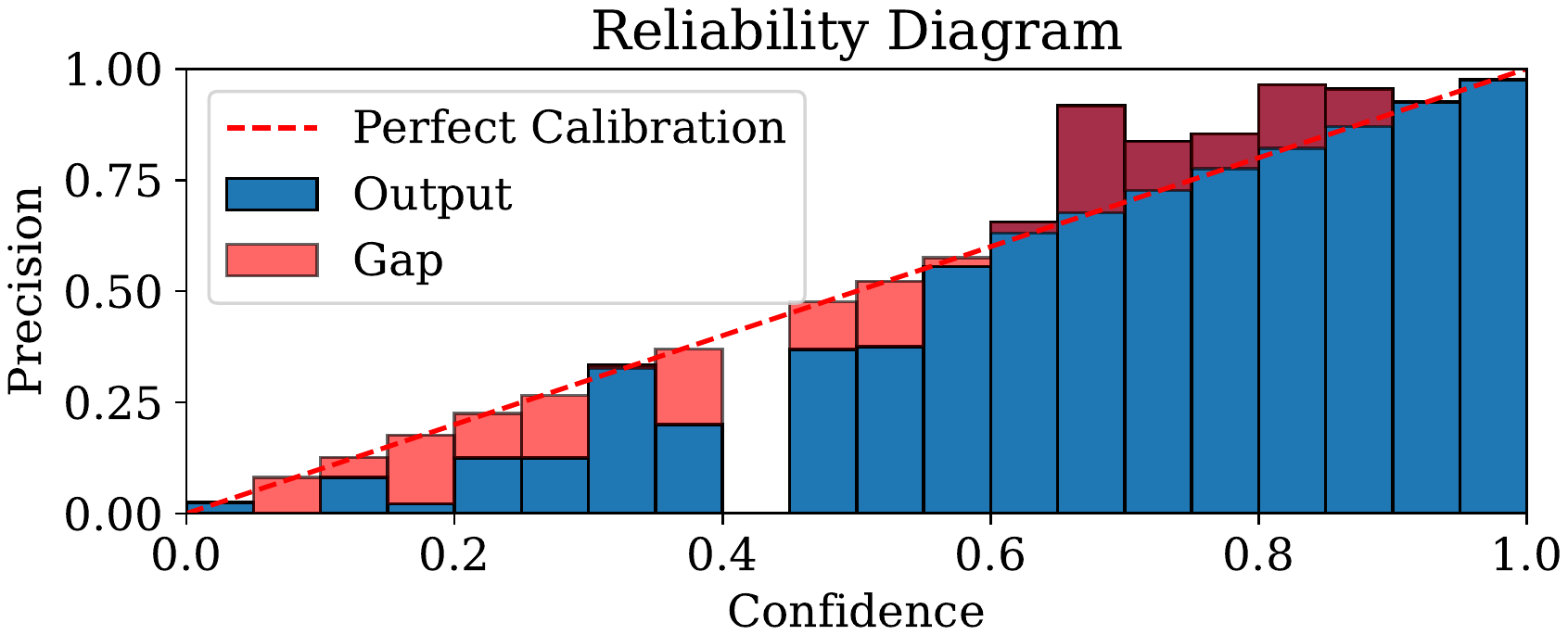}  
    \caption{VOC - Ours}
    \label{fig:VOC_ours}
\end{subfigure}
\begin{subfigure}{.24\textwidth}
    \centering
    \includegraphics[width=.95\linewidth, height=1.8cm]{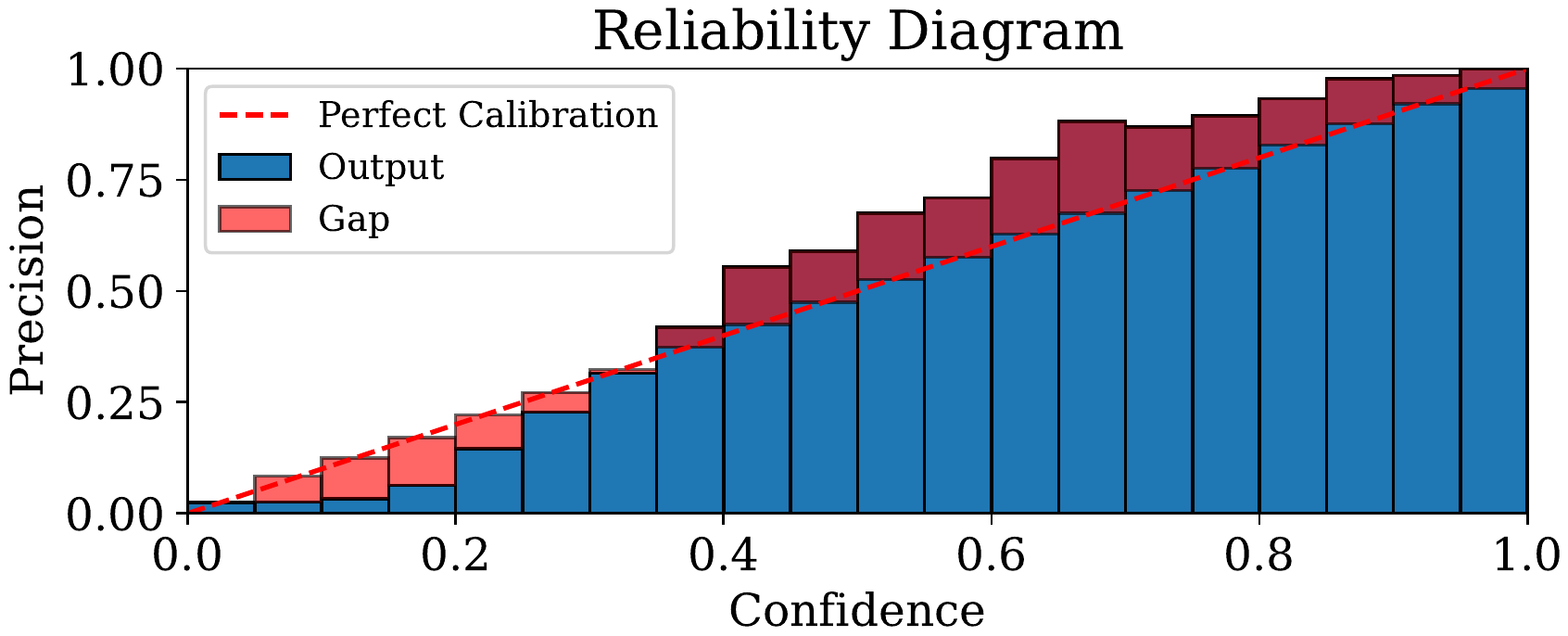}  
    \caption{Sim10K to CS- baseline (FCOS)}
    \label{fig:Sim10k_CS_base}
\end{subfigure}
\begin{subfigure}{.24\textwidth}
    \centering
    \includegraphics[width=.95\linewidth, height=1.8cm]{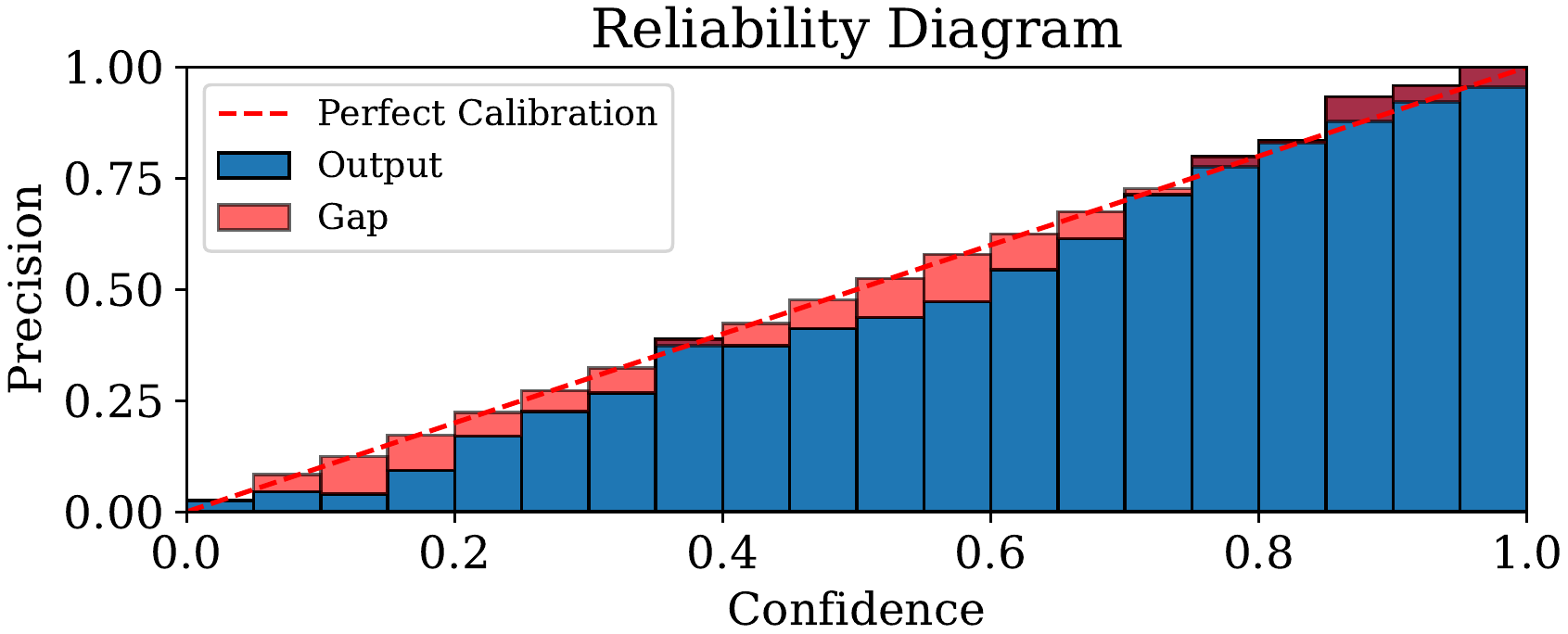}  
    \caption{Sim10K to CS - Ours}
    \label{fig:Sim10k_CS_ours}
\end{subfigure}
\vspace{-0.5em}
\caption{Reliability diagrams for baseline and our method.}
\label{fig:reliability_diag}
\vspace*{-\baselineskip}
\end{figure*}

\noindent \textbf{In-domain experiments:}
We compare the in-domain performance on five challenging datasets with the models trained with task-specific loss of FCOS in \Cref{tab:in_domain_perfor}. The results reveal that our train-time calibration method (MCCL) consistently improves the calibration performance of the task-specific losses. Notably, when added to the task-specific loss of FCOS, our MCCL reduces the D-ECE by 5.86\% and 1.76\% in VOC and CS datasets, respectively.

\noindent \textbf{Out-of-domain experiments:}
\Cref{tab:out of domain} and \Cref{tab:domain-voc} report out-of-domain performance on eight challenging shifts. We see that our MCCL is capable of consistently improving the calibration performance in all shift scenarios. We notice a major decrease in D-ECE of 2.91\% in Sim10K to CS shift. Similarly, we observe a reduction in D-ECE by a visible margin of 2.47\% for CS to CS-foggy (CS-F).


\noindent \textbf{Comparison with post-hoc method:} We choose temperature scaling (TS) as post-hoc calibration for comparison. The temperature parameter T is optimized using a hold-out validation set to re-scale the logits of the trained model (FCOS). \Cref{tab:component_analysis} compares the performance of TS with our method (MCCL) on COCO, Sim10K, CS and COCO corrupted datasets. We note that TS performs inferior to our method and to baseline. This could be because when there are multiple dense prediction maps, as in FCOS, it is likely that a single temperature parameter T will not be optimal for the corresponding logit vectors.


\noindent \textbf{Test accuracy/precision:} We note that in addition to consistently reducing D-ECE, our MCCL also preserves the mAP or AP@0.5 in almost all cases. In the in-domain experiments (\Cref{tab:in_domain_perfor}), the maximum reduction in AP@0.5 is only 0.98\% in the Sim10K dataset. 
%
In the out-of-domain experiments (\Cref{tab:out of domain} \& \Cref{tab:domain-voc}), it mostly remains same in KITTI to CS, CS to BDD100K, VOC to watercolor, and VOC to comic shifts.


 
\noindent \textbf{Overcoming under/overconfidence:} We plot confidence histogram (\Cref{fig:confidence_histogram}) and reliability diagrams (\Cref{fig:reliability_diag}) to illustrate the effectiveness of our method in mitigating overconfidence or underconfidence. In confidence histograms (\Cref{fig:confidence_histogram}) from Sim10K in-domain and CS to CS-F out-of-domain datasets, the average confidence is greater than the average precision which
indicates the overconfident model. Our method reduces this gap in both scenarios compared to the baseline (FCOS) method and alleviates the overconfidence of the baseline.
Similarly, the reliability diagrams (\Cref{fig:reliability_diag}) for VOC in-domain and Sim10K to CS domain shifts reveal that our method can mitigate both underconfident and overconfident predictions by a visible margin.

\begin{table}[!htp]
 \tabcolsep=0.09cm
 \centering
 \scalebox{0.86}{
\begin{tabular}{lllll}
  \toprule
  \multicolumn{5}{c}{Comparison with Deformable DeTR (Baseline)}\\
  \midrule
  Dataset    &  \multicolumn{2}{c}{Baseline}  & \multicolumn{2}{c}{Ours}  \\
       & D-ECE  & AP@0.5  & D-ECE  & AP@0.5 \\
  
  \midrule
       
    Sim10K  & 7.51  & 89.85 & \textbf{6.36}  & 89.63 \\
    KITTI  & 6.31  & 96.76 & \textbf{3.87} & 96.55 \\
    CS     & 9.74  & 71.03  & \textbf{9.69}  & 70.63 \\
    COCO  & 7.92  & 62.93 & \textbf{7.89}  & 62.95 \\
    VOC   & 6.13  & 66.83  & \textbf{5.65}  & 65.67 \\
 \midrule
 
    Sim10K $\,\to\,$ CS  & 7.28  & 49.65 & \textbf{6.79}  & 51.55 \\
    KITTI $\,\to\,$ CS  & 12.93  & 29.93  & \textbf{12.57}  & 29.62 \\
    CS $\,\to\,$ CS-F     & 9.54  & 25.73  & \textbf{9.39}  & 25.39 \\
    COCO $\,\to\,$ Cor-COCO  & 6.77  & 35.51 & \textbf{6.71}  & 35.02 \\
    VOC $\,\to\,$ Clipart   & 4.32  & 16.40  & \textbf{3.55}  & 18.11 \\
    VOC $\,\to\,$ Watercolor   & 6.60  & 27.23  & \textbf{6.56}  & 26.51 \\
    VOC $\,\to\,$ Comic   & 6.49  & 9.41  & \textbf{6.31}  & 9.54 \\
  
  \bottomrule
  
\end{tabular}
}
\caption{\label{tab:DETR_in_out_perf} Comparison of calibration performance of models trained with ViT-based object detector (Deformable DeTR) method and after integrating our method (MCCL).} 
\vspace*{-\baselineskip}
\end{table}

  
       
 
  
  

\begin{figure}[!ht]
  \centering
  \includegraphics[width=\linewidth]{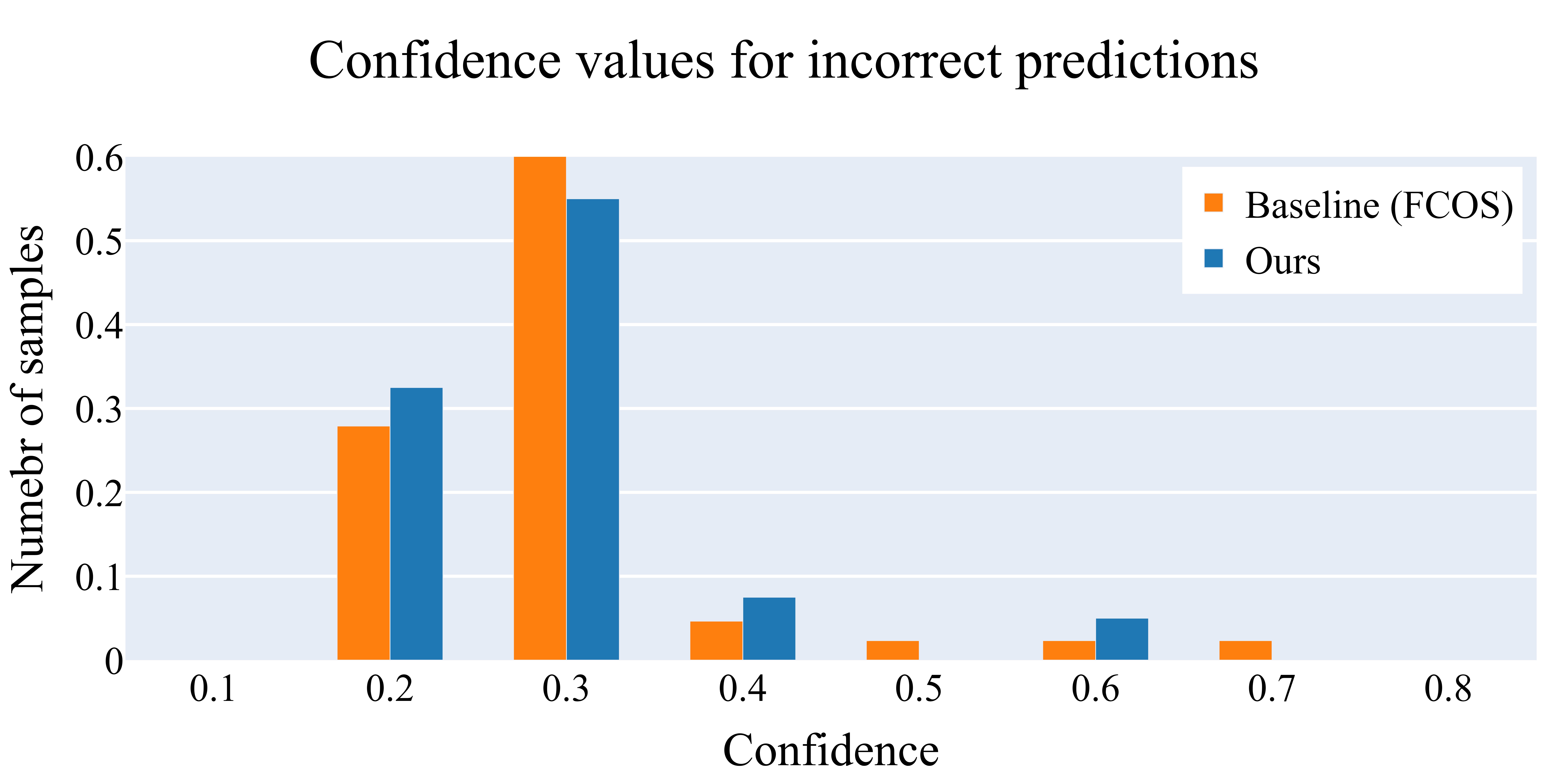}
\vspace{-1em}
\caption{Histogram of confidence values for incorrect predictions by baseline (FCOS) and our method in COCO dataset.}
   \label{fig:histogram}
   \vspace*{-\baselineskip}
\end{figure}

\noindent \textbf{Confidence values of incorrect detections:}
We analyse the confidence of our method in case of incorrect predictions (\Cref{fig:histogram}). Compared to baseline, our method is capable of reducing the confidence of incorrect predictions over the whole spectrum of confidence range. 

\noindent \textbf{With another baseline:} \Cref{tab:DETR_in_out_perf} reports results with ViT-based object detector, namely Deformable DETR \cite{zhu2021deformable}. Compared to FCOS, the Deformable DETR, is already a relatively strong baseline in calibration error. We observe that our MCCL reduces the calibration error (D-ECE) for both in-domain and out-of-domain predictions. The major improvement (2.44\% reduction in D-ECE) in calibration performance is observable for KITTI in-domain predictions.


\subsection{Ablation study}
\label{subsection:Ablation study}

\begin{table*}[!htp]
\centering
\tabcolsep=0.25cm
\scalebox{0.93}{
\begin{tabular}{lllllllll}
  \toprule
  \multicolumn{9}{c}{Ablation study}\\
  \midrule
  Methods   &  \multicolumn{2}{c}{COCO$\,\to\,$COCO}  & 
  \multicolumn{2}{c}{Sim10K$\,\to\,$ Sim10K}  & \multicolumn{2}{c}{Sim10K$\,\to\,$ CS} & \multicolumn{2}{c}{COCO$\,\to\,$ Cor-COCO}  \\
 
       & D-ECE  & AP@0.5  & D-ECE  & AP@0.5 & D-ECE  & AP@0.5 & D-ECE  & AP@0.5\\
  
  \midrule
    Baseline (FCOS)    & 15.42  & 54.91  & 12.90  & 88.01 & 9.51  & 45.18 & 15.90   &  30.01\\
    Post-hoc (TS) & 17.34  &54.77  & 17.99  & 29.97 &14.66 & 87.29 &24.03 &45.91\\
     AvUC ~\cite{krishnan2020improving}     & 15.17   &  54.73   & 13.24  & 87.91  & 10.64   &  39.39 & 15.78   &  29.75\\
	 
     MDCA ~\cite{hebbalaguppe2022stitch}   & 15.25  & 54.44  & 13.99  & 87.95 & 11.14  & 39.51 & 15.33   &  30.07\\

    Ours (w/o $\mathcal{L}_{LC}$ \& $\mathcal{L}_{MCC}$) & 15.26 & 54.25 & \underline{13.12} & 86.34 & 8.87 & 42.89  & 15.28  &  29.67\\
     Ours (w/o $\mathcal{L}_{LC}$)    & \underline{15.00}  & 54.11 & 13.00  & 87.86 & \underline{8.63}  & 45.31  & \underline{14.52}   &  29.73\\
     Ours (w/o $\mathcal{L}_{MCC}$) & 15.12 & 54.40 & \underline{12.86} & 87.24 & 9.12 & 41.14  & 15.54  &  29.61\\
     Ours (MCCL)    & \textbf{14.94}  & 54.85  & 
     \textbf{11.18}  & 86.47 & 
     \textbf{6.60}  & 44.29 & \textbf{14.45}   &  29.96\\
  
  \bottomrule
  
\end{tabular}}
\vspace{-0.1em}
\caption{\label{tab:component_analysis} Ablations in MCCL and comparison of MCCL with TS, and classification-based train-time losses: MDCA\cite{hebbalaguppe2022stitch} and AvUC\cite{krishnan2020improving}.}

\end{table*}

 
  
	 
    
  
  


 \begin{figure}[!htp]
  \centering
  \includegraphics[trim={0 0 0 2cm},clip, width=\linewidth]{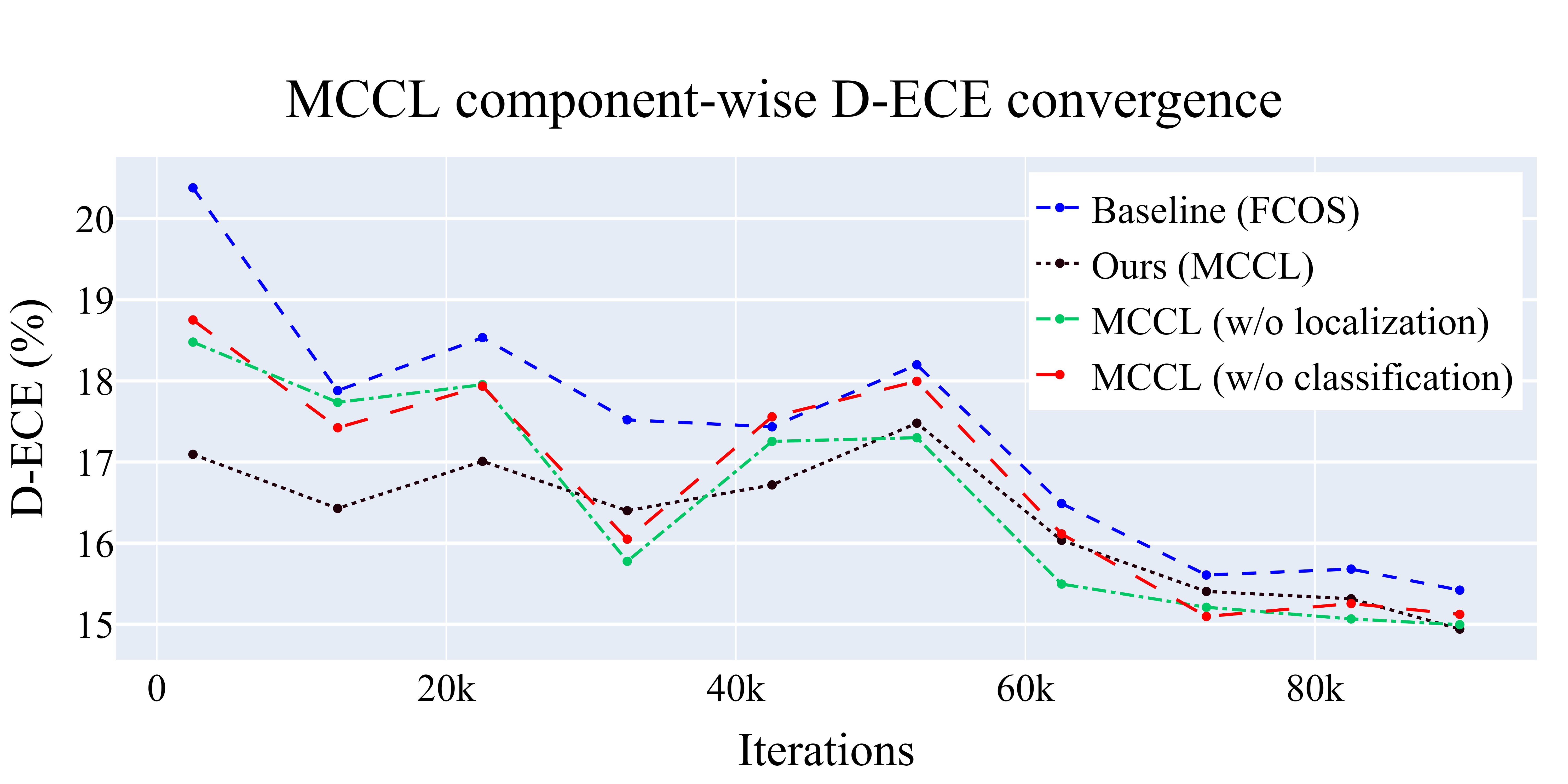}
  
  \vspace{-0.5em}

   \caption{D-ECE convergence for baseline, the classification and localization components of our MCCL, and MCCL.}
   
   \label{fig:coco-covergence}
   \vspace*{-\baselineskip}
\end{figure}

\begin{figure*}[!htp]
\centering
\begin{subfigure}{.24\textwidth}
    \centering
    \includegraphics[width=.95\linewidth]{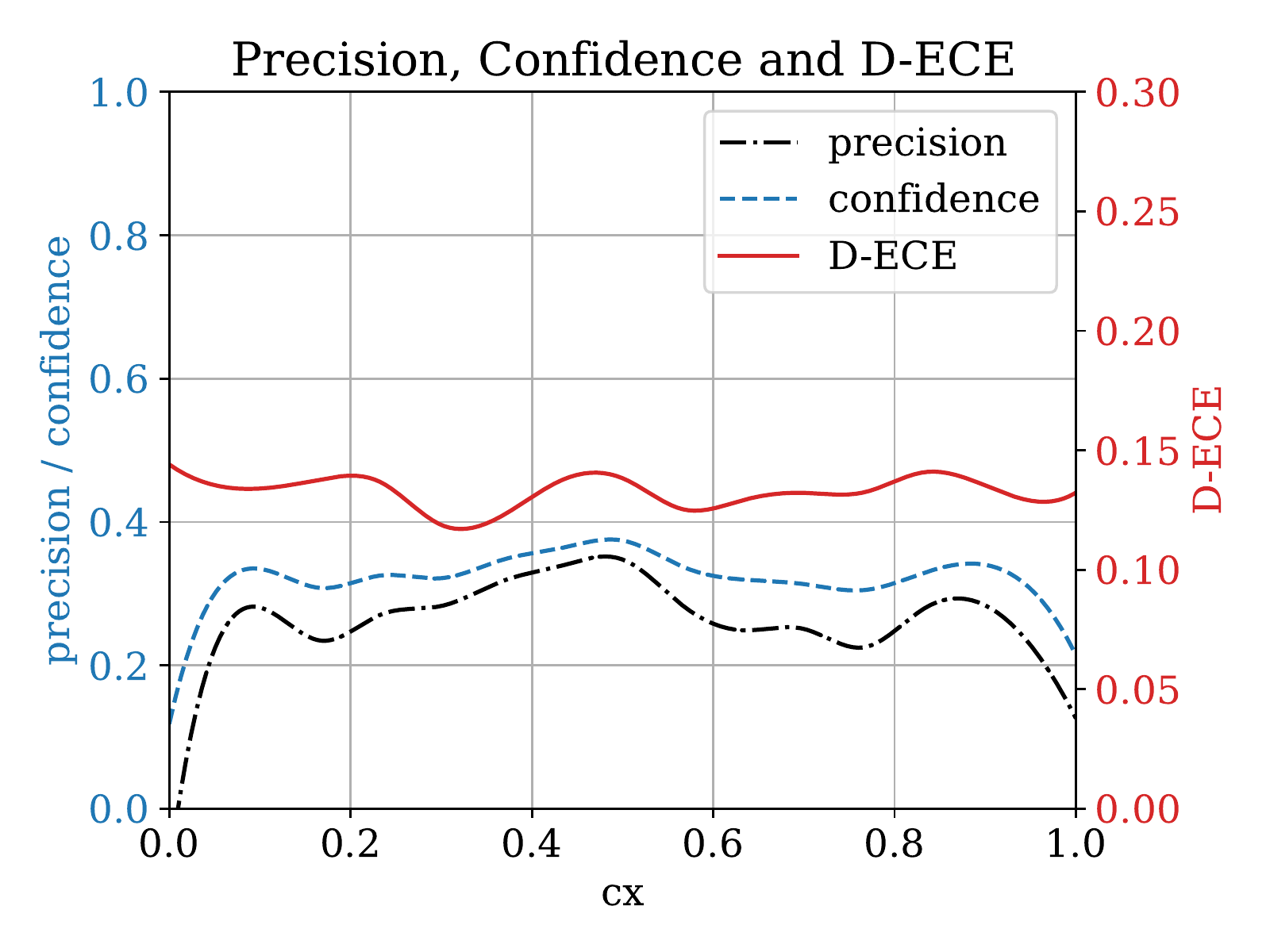}  
    \caption{Baseline $c_{x}$}
    \label{fig:base_cx}
\end{subfigure}
\begin{subfigure}{.24\textwidth}
    \centering
    \includegraphics[width=.95\linewidth]{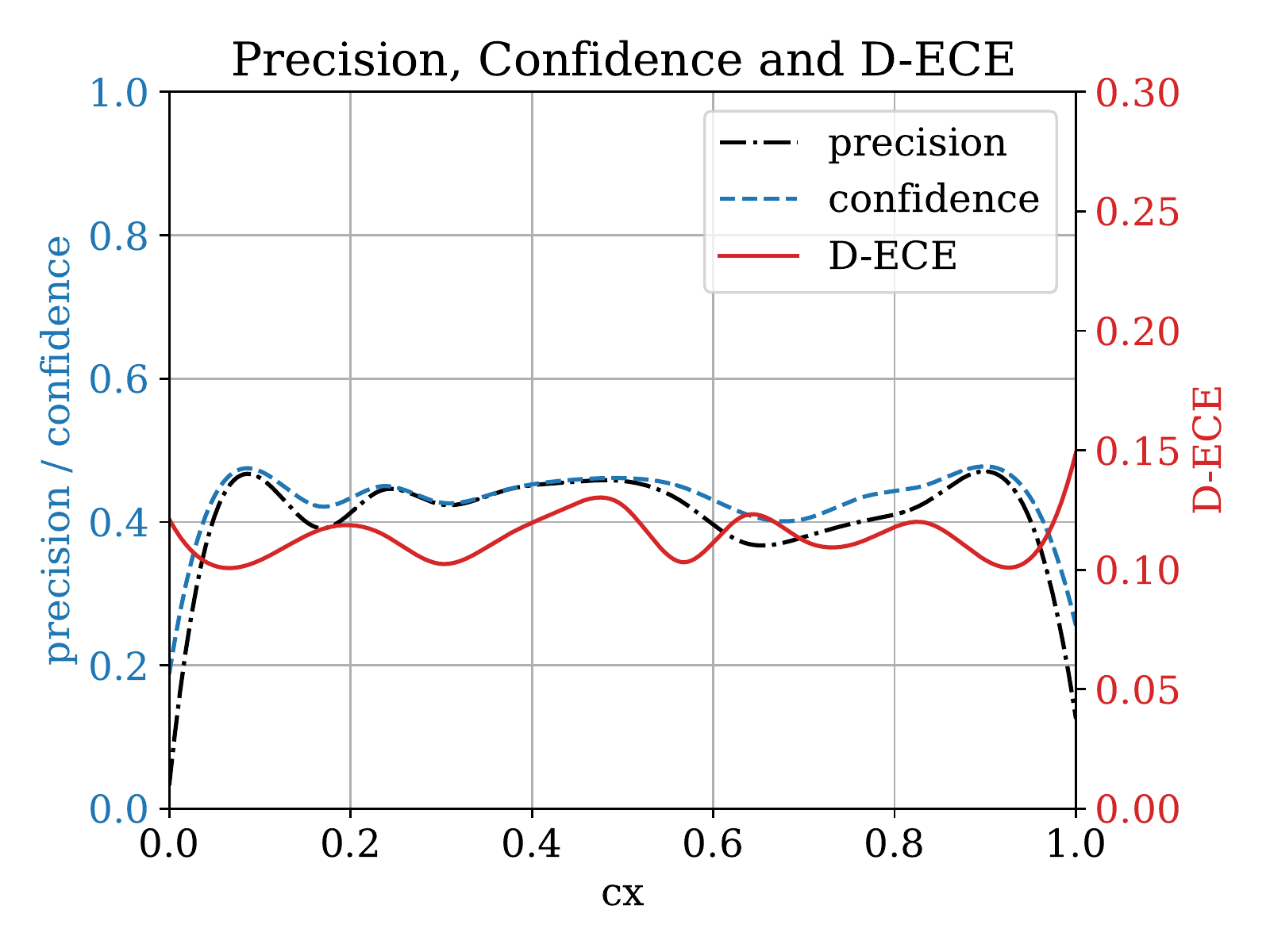}  
    \caption{Ours $c_{x}$}
    \label{fig:ours_cx}
\end{subfigure}
\begin{subfigure}{.24\textwidth}
    \centering
    \includegraphics[width=.95\linewidth]{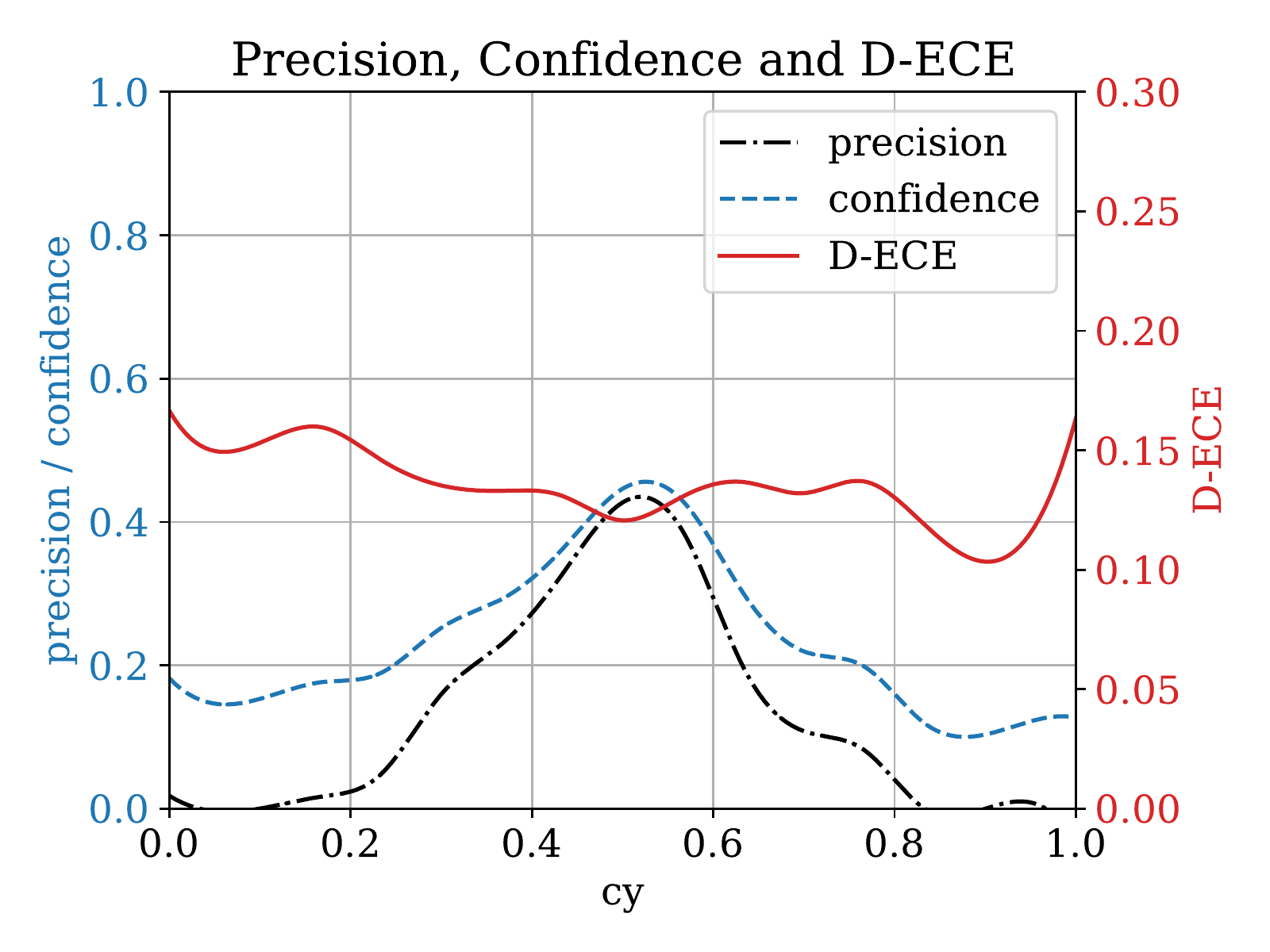}  
    \caption{Baseline $c_{y}$}
    \label{fig:base_cy}
\end{subfigure}
\begin{subfigure}{.24\textwidth}
    \centering
    \includegraphics[width=.95\linewidth]{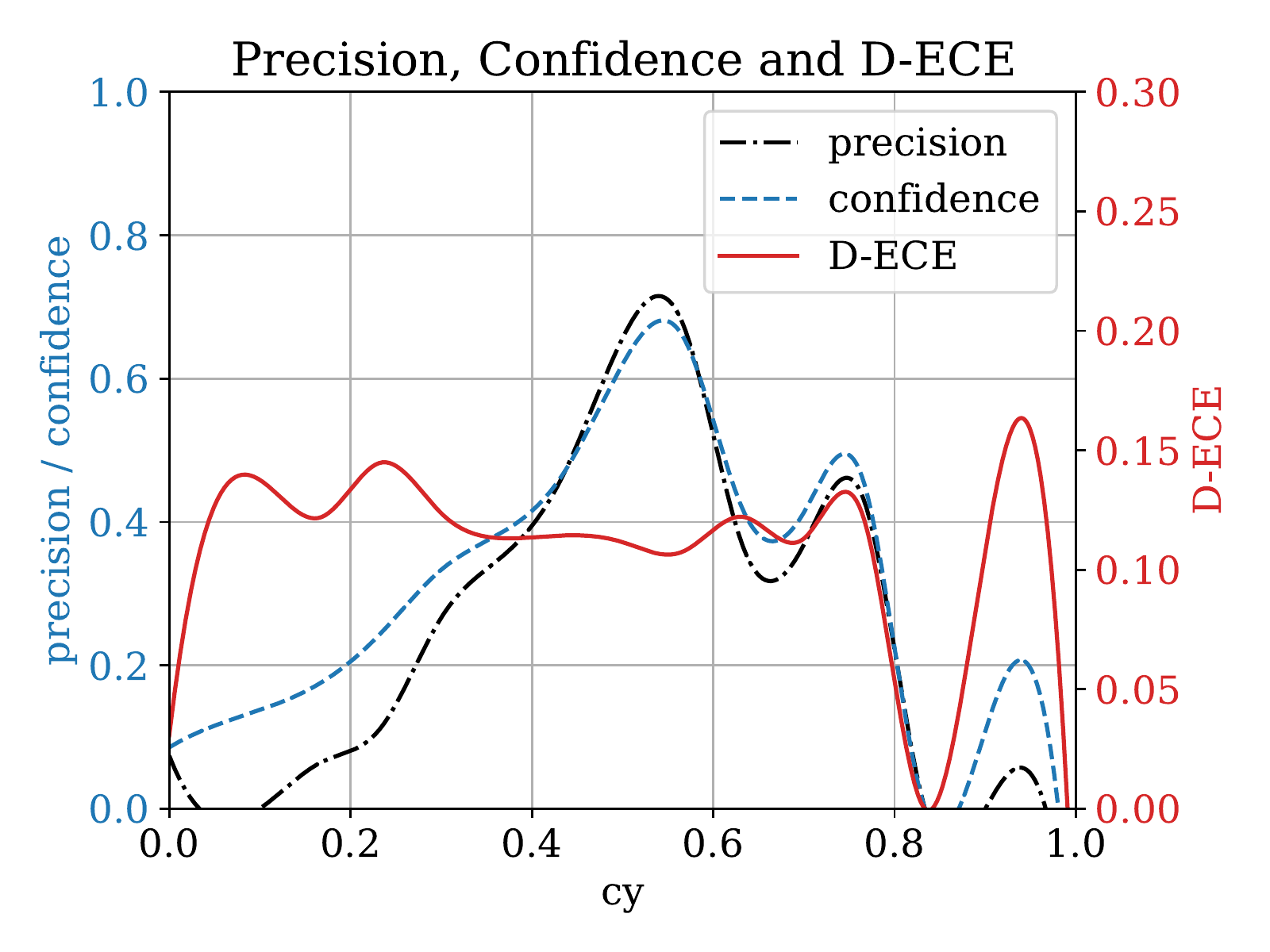}  
    \caption{Ours $c_{y}$}
    \label{fig:ours_cy}
\end{subfigure}
\vspace{-0.5em}
\caption{Calibration precision, confidence, and ECE with IOU @0.5 of (a) baseline (FCOS) relative to center-x ($c_{x}$), (b) our method relative to center-x ($c_{x}$) (c) baseline (FCOS) relative to center-y ($c_{y}$), and (d) our method relative to center-y ($c_{y}$).}
\label{fig:accuracy_confidence_ece}
\end{figure*}

\begin{figure*}[!htp]
\centering

\begin{subfigure}{.24\textwidth}
    \centering
    \includegraphics[width=\linewidth]{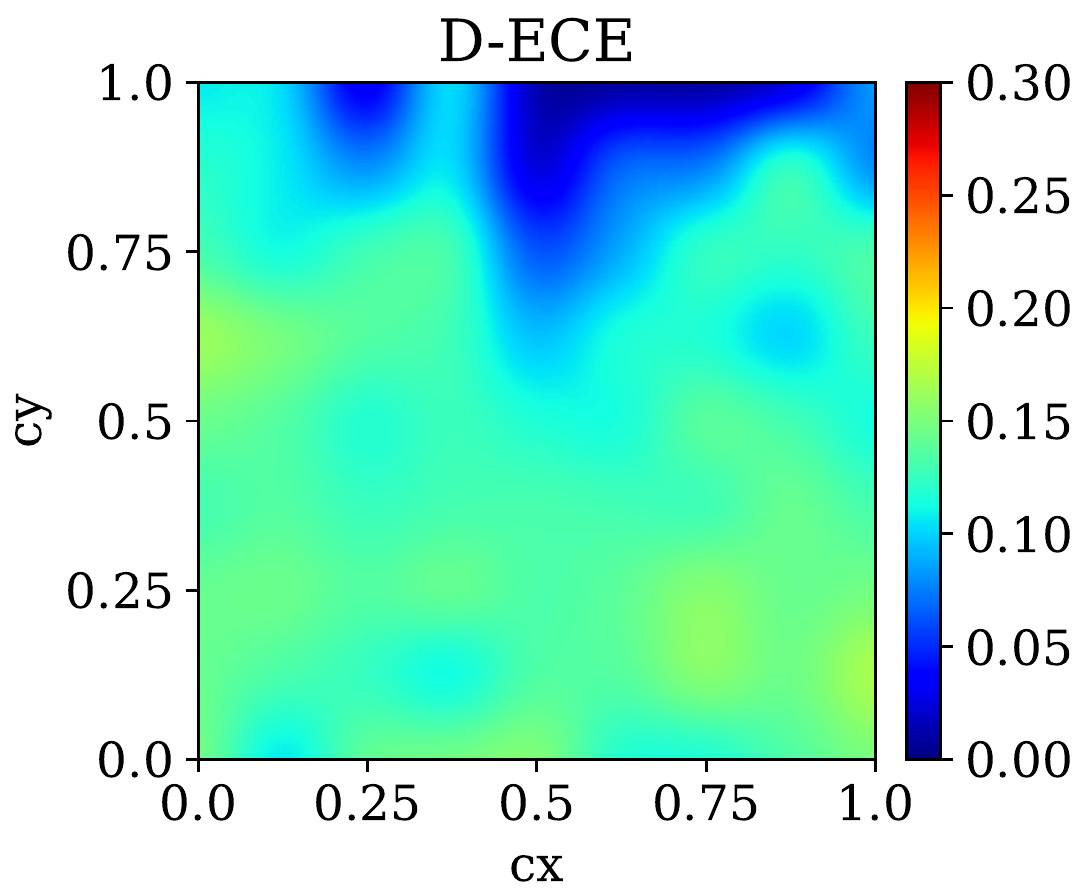}  
    \caption{Baseline $c_{x}$ \& $c_{y}$}
    \label{fig:base_cx_cy}
\end{subfigure}
\begin{subfigure}{.24\textwidth}
    \centering
    \includegraphics[width=\linewidth]{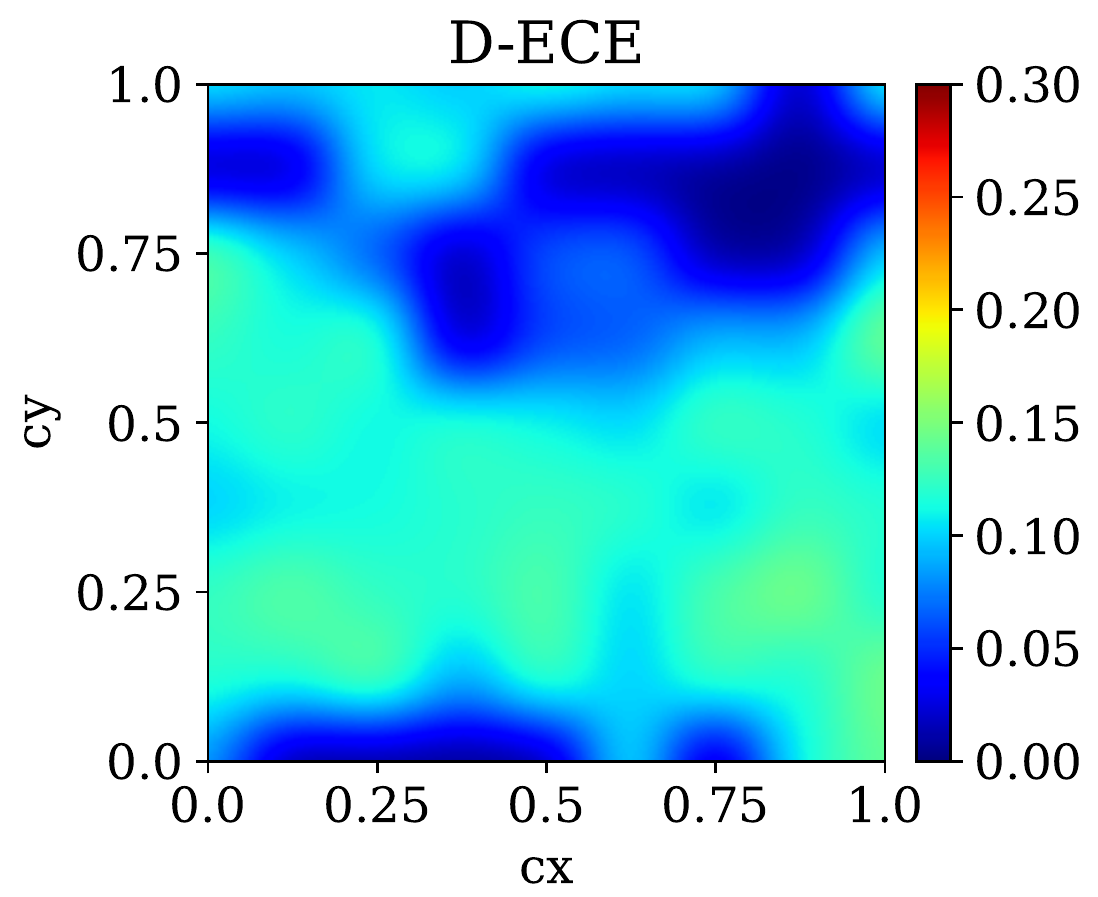}  
    \caption{Ours $c_{x}$ \& $c_{y}$}
    \label{fig:ours_cx_cy}
\end{subfigure}
\begin{subfigure}{.24\textwidth}
    \centering
    \includegraphics[width=\linewidth]{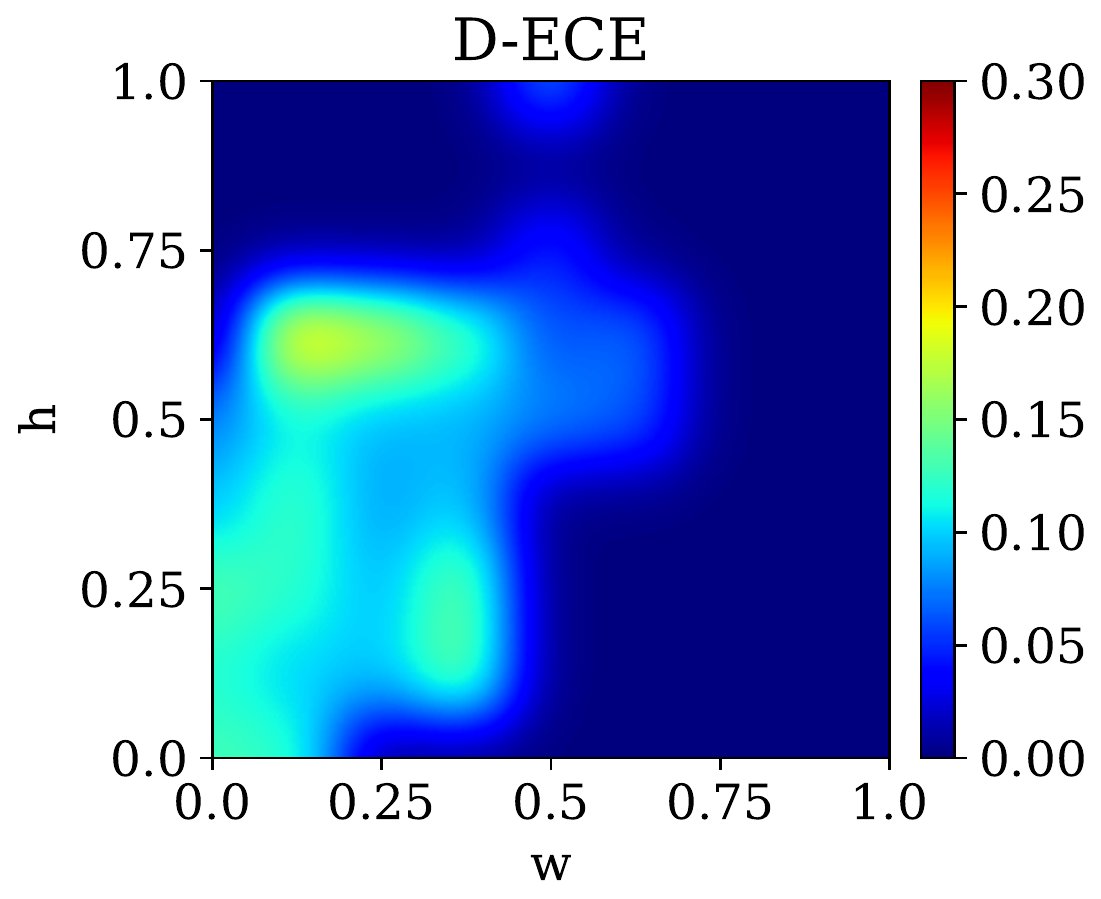}  
    \caption{Baseline   to w \& h}
    \label{fig:base_w_h}
\end{subfigure}
\begin{subfigure}{.24\textwidth}
    \centering
    \includegraphics[width=\linewidth]{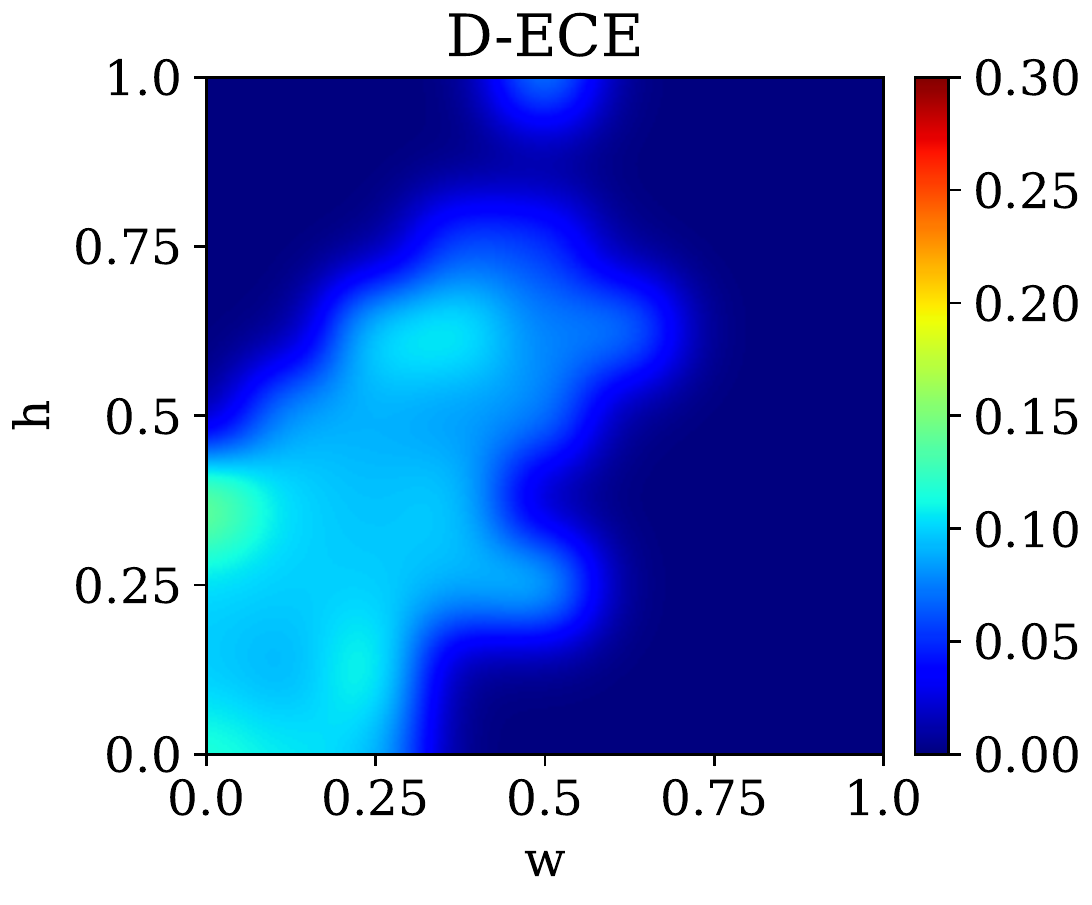}  
    \caption{Ours w \& h}
    \label{fig:our_w_h}
\end{subfigure}
\vspace{-0.5em}
\caption{Calibration heatmap of (a,b) baseline (FCOS) and our method over center-x ($c_{x}$) and center-y ($c_{y}$) with IOU @0.5, (c,d) baseline (FCOS) and our method over width (w) and height (h) with IOU @0.5.}
\label{fig:heat_maps}
\end{figure*}

\noindent \textbf{Impact of each component in MCCL:} We report the result of ablation experiments for validating the performance contribution of different components in our method (MCCL) (\Cref{tab:component_analysis}). Moreover, we report the calibration performance of two train-time calibration losses for image classification: MDCA~\cite{hebbalaguppe2022stitch} and AvUC ~\cite{krishnan2020improving}. We can observe the following trends from \Cref{tab:component_analysis}. The calibration performance of our MCCL is not due to providing only the class-wise mean logits and mean bounding box parameters to classification loss and regression loss of detection-specific loss, respectively, (ours w/o $\mathcal{L}_{LC}$ \& $\mathcal{L}_{MCC}$). Both $\mathcal{L}_{MCC}$ and $\mathcal{L}_{LC}$ are integral components of our method (MCCL). They are complementary to each other and their proposed combination is vital to delivering the best calibration performance. For instance, in Sim10K to CS shift, the proposed combination of $\mathcal{L}_{MCC}$ and $\mathcal{L}_{LC}$ achieves a significant reduction in D-ECE compared to MCC and LC alone. Further, the classification-based calibration losses are sub-optimal for calibrating object detection methods.



\noindent \textbf{D-ECE convergence:} \Cref{fig:coco-covergence} compares the convergence of D-ECE  for baseline, the two components (classification and localization) of our method (MCCL), and MCCL. Although our MCCL and its two constituents does not directly optimize the D-ECE metric, they provide improved D-ECE convergence compared to the baseline.

\noindent \textbf{Impact on location-dependent calibration:} \Cref{fig:accuracy_confidence_ece} and \Cref{fig:heat_maps} depict that miscalibration error (D-ECE) relies highly on the relative object location ($c_x$, $c_y$) and/or its relative width and height ($w$, $h$). Moreover, it tends to increase as we approach image boundaries. \Cref{fig:accuracy_confidence_ece} plots the precision, confidence and D-ECE over individual parameters i.e.~$c_x$. \Cref{fig:heat_maps} plots 2D calibration heatmaps over object location and width/height, where each location in a heatmap represents D-ECE. Both figures show that, compared to baseline, besides other locations, our MCCL can decrease D-ECE at image boundaries. \Cref{fig:accuracy_confidence_ece} also shows that, compared to baseline, our MCCL allows the adaptation of confidence score at all image locations differently by adjusting the shape of confidence curve accordingly.


\noindent \textbf{MCDO overhead \& its Tradeoff analysis:} Table~\ref{tab:MCDO_overhead} reveals that, in our implementation, upon increasing Monte-Carlo dropout (MCDO) passes N=$\{3,5,10,15\}$, there is a \emph{little overhead} in time cost over N=1. Table~\ref{tab:mc_effect} shows the impact of varying the number of MC dropout passes (N) on calibration performance. Upon increasing the N, we see improved calibration, especially in OOD scenario.

\begin{table}[!htp]
\centering
\resizebox{0.46\textwidth}{!}{
\begin{tabular}{llllll}
  \toprule
  \multicolumn{6}{c}{Time cost with MC dropout}\\
  \midrule
  N    &   N=1 & N=3 & N=5 & N=10 & N=15  \\
  \midrule
  Time per iteration(s) & 0.143 & 0.197 & 0.243 & 0.350 & 0.463 \\ 
  Increment  & - & 0.372 & 0.694 & 1.442 & 2.230 \\

\bottomrule
  
\end{tabular}}
\vspace{-0.0em}
\caption{Increment in time per iteration (in secs) over N=1 upon increasing MC forward passes.}\label{tab:MCDO_overhead}
\vspace*{-1em}   
\end{table}


\begin{table}[htp]
\centering
\tabcolsep=0.09cm
 \resizebox{0.35\textwidth}{!}{
\begin{tabular}{lllllll}
  \toprule
  \multicolumn{7}{c}{Calibration performance with MC dropout}\\
  \midrule
    Method & Metric & baseline & N=3 & N=5 & N=10 & N=15\\
  
  \midrule
      \multirow{2}{3em}{COCO}
      & D-ECE & 16.57 & 16.27 & 16.25 & 16.19 & 16.12 \\
     & AP@0.5 & 52.34 & 51.00 & 51.49 & 51.23 & 51.89 \\
     \midrule
    \multirow{2}{3em}{COCO-Corr.}
      & D-ECE & 17.27 & 16.18 & 16.21 & 15.87 & 16.17 \\
     & AP@0.5 &  29.25 & 28.24 & 28.59 & 28.31 & 28.80  \\
  
  \bottomrule
\end{tabular}}
\vspace{-0.0em}
\caption{\small Impact on D-ECE after increasing N in MC dropout.}
\vspace*{-\baselineskip}
\label{tab:mc_effect}
\end{table}

\section{Conclusion}

Very little to no attempts have been made towards studying the calibration of object detectors. In this paper, we explored this direction and presented a new train-time technique for calibrating DNN-based object detection methods. At the core of our method is an auxiliary loss which aims at jointly calibrating multiclass confidence and box localization after leveraging their predictive uncertainties. Extensive experiments reveal that our method can consistently reduce the calibration error of object detectors from two different DNN-based object detection paradigms for both in-domain and out-of-domain detections.


{\small
\bibliographystyle{ieee_fullname}
\bibliography{egbib}
}

\end{document}